\documentclass[10pt,journal]{IEEEtran}
\usepackage[utf8]{inputenc}
\usepackage{multirow}
\usepackage{xcolor}

\usepackage{cite}

\ifCLASSINFOpdf
\else
  \usepackage[dvips]{graphicx}
  \graphicspath{{./pictures/}}
  \DeclareGraphicsExtensions{.eps}
\fi
\usepackage{amsmath}
\usepackage{array}

\ifCLASSOPTIONcompsoc
  \usepackage[caption=false,font=normalsize,labelfont=sf,textfont=sf]{subfig}
\else
  \usepackage[caption=false,font=footnotesize]{subfig}
\fi
\usepackage{url}

\begin{document}
%
\normalsize\title{{Learning to automatically catch Potholes in worldwide road scene images}}

\author{J. Javier Yebes, David Montero and Ignacio Arriola
  \thanks{J. Javier Yebes, David Montero and Ignacio Arriola are with the Department of Intelligent Transport Systems 
  and Engineering, Vicomtech, Paseo Mikeletegi 57, 20009 Donostia/San Sebastián, Spain (e-mail: jyebes; dmontero; iarriola@vicomtech.org).}
  \thanks{© 2021 IEEE. Personal use of this material is permitted. Permission from IEEE must be obtained for all other uses, 
  in any current or future media, including reprinting/republishing this material for advertising or promotional purposes, 
  creating new collective works, for resale or redistribution to servers or lists, or reuse of any copyrighted component of this work in other works.}
}

\markboth{IEEE ITS Magazine}%
{Learning to automatically catch Potholes in worldwide road scene images}
%



\IEEEtitleabstractindextext{%
\begin{abstract}
Among several road hazards that are present in any paved way in the world, potholes are one of the most annoying and involving higher maintenance costs. There is an increasing interest on the automated detection of these hazards enabled by technological and research progress. Our work tackled the challenge of pothole detection from images of real world road scenes. The main novelty resides on the application of latest progress in Artificial Intelligence to learn the visual appearance of potholes. We built a large dataset of images with pothole annotations. They contained road scenes from different cities in the world, taken with different cameras, vehicles and viewpoints under varied environmental conditions. Then, we fine-tuned four different object detection models based on Deep Neural Networks. We achieved mean average precision above 75\% and we used the pothole detector on the Nvidia DrivePX2 platform running at 5-6 frames per second. Moreover, it was deployed on a real vehicle driving at speeds below 60 km/h to notify the detected potholes to a given Internet of Things platform as part of AUTOPILOT H2020 project.
\end{abstract}

}

\maketitle

\IEEEdisplaynontitleabstractindextext

%
\IEEEpeerreviewmaketitle

\section{Introduction}\label{sec:intro}

\IEEEPARstart{R}{oad defects} are inherent to any road in the world due to several reasons such as weather, high traffic load and heavy vehicles. In some cases, large investments in infrastructure were done a long time ago and road surfaces have become more prone to deterioration such that they require frequent inspection and maintenance~\cite{ref:MaedaArxiv2018}. Other places around the world present different limitations and environmental conditions that influence the quality and state of the pavement~\cite{ref:Madli2015Sensors}\cite{ref:NienaberSUN2015}. As a matter of fact, the road network is a valuable asset in any country because it serves the increasing demand of transportation for goods and people~\cite{ref:GlobalTransport2050}. Consequently, an important budget is reserved for road reconstruction. For instance, Spain assigns an averaged 60\% of its road maintenance budget to surface rehabilitation~\cite{ref:FomentoES}.

Commonly, road inspection has been carried out by qualified maintenance staff who drives along the road network, also stopping at several locations, to monitor and report encountered road hazards. However, automating these monitoring tasks can improve the safety of the staff and the effective detection of the hazards. In the recent years, technological progress has seen the installation of sensors and specialized equipment in maintenance vans, e.g. for the detection of road cracks~\cite{ref:GavilanSensors2011}. Additionally, the progress in ADAS systems and the large investment in autonomous driving technologies have enabled the integration of multiple sensors in cars and the multi-modal perception of the environment~\cite{ref:Janai17Arxiv}. In parallel, the advances in Artificial Intelligence (AI)~\cite{ref:Duda2010PR}~\cite{ref:Krizhevsky2012NIPS} have procured an effective use of the collected data from the sensors.

Among several types of road hazards (cracks, rutting, deteriorated markings, etc.) the target of our work is the automated detection of road potholes. They are present in worldwide roads causing discomfort to drivers, damages to vehicles and non-negligible repairing costs to public and private roads. For instance, the Department of Transport in UK stated in 2014 that more than \pounds3 billion were spent nationally on road repairs. Also, the Royal Automotive Club (RAC) in UK estimated that vehicle repairing costs were around \pounds100 million for all affected motorists and general surveys estimated that American drivers pay an average annual cost of \textdollar300 each to fix car damage due to potholes. 
Besides, specific funding has been provided to research on durable pothole repairs~\cite{ref:POTHOLE}. In fact, recent harsh winters and springs, weather changes and transportation demands are causing a rapid increase of road potholes.

Our research has received funds from the AUTOPILOT H2020 project~\cite{ref:AutopilotWebsite}, which will deploy, test and demonstrate automated services based on Internet of Things (IoT) in five driving modes. In the Highway Pilot use case, a cloud service merges the sensors' measurements from different IoT devices to locate and characterize road hazards. The goal is to provide the following vehicles with meaningful warnings and driving recommendations to manage the hazards in a safer or more pleasant way. For a better understanding of the scenario, we assume the following:
Firstly, a vehicle equipped with different systems has the role of IoT device, which is comparable to smartphones and other wearables that can send/receive messages to/from IoT platforms. Secondly, in-vehicle systems include AI modules that process data and produce low-bandwidth messages that are wirelessly sent to IoT platforms.

Within this background, our main research motivation is the automated visual detection of road potholes from a frontal colour camera on board a vehicle. Once potholes are detected, their location is reported to a given IoT platform. Current state of the art is predominantly based either on sensing potholes with accelerometers~\cite{ref:BhattArxiv2017} or cameras~\cite{ref:NienaberSUN2015}. Accelerometer-based detection requires that the vehicle drives over the potholes, which is usually not the case as the driver will try to avoid them. Vision-based detection is naturally seen as the same process in which drivers perceive the environment and anticipate to possible road hazards. For the latter one, classical image processing and machine learning approaches have been evaluated on images of certain world regions. Our strategy is to automatically learn the visual appearance of worldwide potholes using latest advances in AI, i.e. Deep Neural Networks. 

Therefore, the contributions of this paper are the following:
\begin{itemize}
\item We \textbf{built a dataset} with manual annotations from several \textbf{places around the world} that include images from Europe, America, Asia and Africa. It is composed of challenging scenes captured from different cameras, viewpoints and under varied environmental conditions.
\item We fine-tuned and evaluated 4 different Deep Neural Networks \textbf{(DNNs)} for the \textbf{detection of road potholes on images}. We achieved high detection ratios ($mAP>75\%$) considering the high intraclass variance.
\item The pothole detector was \textbf{tested on the Nvidia DrivePX2} platform for embedding ADAS in driverless vehicles.
\item As part of AUTOPILOT project, the pothole detector was \textbf{integrated on a real vehicle}. The set-up included an automotive grade camera, General-Purpose computing on Graphics Processing Units (GPGPU) and IoT communication modules on board the vehicle.
\end{itemize}

The remaining of the paper is organized as follows: In Section~\ref{sec:sota} we review sensing modalities and literature related to the detection of road potholes. Next, Section~\ref{sec:aviot} presents the AUTOPILOT European project and related research areas. Section~\ref{sec:learnpotholes} describes our approach to learn the visual appearance of potholes with Deep Neural Networks and Section~\ref{sec:results} explains the created dataset, the experiments carried out and the obtained results. Finally, Section~\ref{sec:conclusion} concludes the paper.

\section{Related work}\label{sec:sota}

The inspection of roads for surface damage is a worldwide challenge. The literature reviewed in this section includes systems from places all around the globe. Commonly, qualified maintenance staff monitors and reports encountered road hazards. In the recent years, technological progress has enabled the installation of specialized equipment in maintenance vans~\cite{ref:GavilanSensors2011}. Nowadays, smartphones are readily available and they can be used to take images on the road from any vehicle~\cite{ref:MaedaArxiv2018}\cite{ref:Lee2018ICCE}.

The remaining of this section is dedicated to detection approaches for our targeted hazards: the road potholes. Attending to sensor modality, there are four categories: \emph{ultrasonic}, \emph{accelerometer}, \emph{image} and \emph{combination of image and depth}.

\subsection{Ultrasonic-based pothole detectors}\label{ssec:ultrasonic}
The system in~\cite{ref:HegdeICVES2014} consisted on a prototype robot vehicle with one ultrasonic sensor and a Zigbee communication module. It was designed to detect potholes with a minimum depth of 1 inch and to broadcast warning messages to vehicles in its vicinity (100m). However, it was tested as a lab prototype without real data. Another system was attached to a motorbike and tested on Indian roads~\cite{ref:Madli2015Sensors} in which the ultrasonic sensor was used to determine the depth and height of potholes and speed humps. However, motorbikes are easy to manoeuvre and drivers will typically try to avoid driving over the hazards, thus reducing its applicability.

\subsection{Accelerometer-based pothole detectors}\label{ssec:sotaacc}

Recently, many studies have explored accelerometer signals from smartphone sensors and on board vehicle sensors. Mobile sensing was researched in~\cite{ref:MednisDCOSS2011} that addressed the differences between sensors embedded in 4 distinct phones and compared various z-axis thresholding algorithms for the detection of potholes. Similarly, \cite{ref:WangMPE2015} presented a real-time pothole detector in which the signals were normalized to account for smartphone position and attitude. However, a single pothole case study of size $51\times58\times6$cm was used.

Given the limitation of simple heuristics and thresholding, Machine Learning (ML) became into scene. Bhatt et. al.~\cite{ref:BhattArxiv2017} presented an intelligent system in a smartphone where the pothole detections (at 5HZ) were marked in a map. They concluded that Support Vector Machines yielded the best accuracy in the classification of road conditions. Alternatively, the PADS~\cite{ref:RenCS2017} system was based on K-Means, tuned thresholds and a tri-axial accelerometer on board the vehicle (no smartphone). PADS solution was based on IoT, sending detections to a remote server and marking their approximate location in Google Maps. A different approach was introduced by Fox et. al.~\cite{ref:Fox2017TMC} for the aggregation and cloud processing of crowd-sourced data from inertial sensors on board vehicles.

As discussed in \cite{ref:RenCS2017}, the accelerometer-based approaches present several drawbacks. Vehicle cushioning mitigates the vibrations produced by potholes, experiments become biased and the system can be confused due to braking and other road anomalies. Moreover, smartphone-based solutions depend on model and type of mounted accelerometers and their location and orientation in the vehicle. Achieving a self-calibration and a correct alignment of axes is not straight-forward~\cite{ref:Almazan2013IV}. Furthermore, these systems assume that at least one of the vehicle wheels passes over the pothole which may not happen if the driver slightly swerves to avoid them or if the potholes are in the center of the lane.

\subsection{Image-based pothole detectors}\label{ssec:sotavision}
A naturalistic approach is to perceive the road scene with cameras and analyse the images in search of potholes, which mimics the visual inspection of humans. One of the early research works consisted on road condition assessment using hyperspectral aerial imagery~\cite{ref:JengoASPRS2005}. 

More recently, two works applied simple image processing techniques for pothole segmentation.\cite{ref:KimJETCIS2014} collected grayscale images from a camera on board a vehicle and \cite{ref:Akagic2017MIPRO} employed colour images from Google search engine using the keyword "pothole". However, both conducted few tests and obtained non-sufficient evidence of the accuracy.

A key reference was~\cite{ref:NienaberSUN2015}, which enabled an initial analysis of the vision-based pothole detection challenge. They released a publicly available and annotated dataset of varied road potholes taken with a camera attached to the vehicle windscreen, driving at normal speed ($<40 Km/h$) and under various illumination conditions. Further details of the dataset are provided in Section~\ref{sec:results}. The image processing techniques in~\cite{ref:NienaberSUN2015} were limited compared to the state of the art in AI. Although reported results were promising, they depended upon several computer vision filters with parameter tuning. It must be observed that the continuation work~\cite{ref:NienaberSSCI2015} evaluated computer vision algorithms for pothole distance estimation but this task is out of the scope of our paper.

In~\cite{ref:Jang2016IET}, a spatio-temporal saliency map was proposed to detect potholes in road scenes with heavy traffic. Grayscale images from a dashcam were used for the purpose and their visual properties were studied. \cite{ref:Azhar2016CCECE} applied ML methods to classify images in two sets as pothole/non-pothole. They employed HOG features and Naive Bayes classifier. Besides, they combined the approach with normalized graph-cut segmentation to locate pothole regions on the positively detected images. For the evaluation, they relied on a private dataset of limited size and reported an elapsed time of ~0.7s per image containing potholes.

Additionally, some newer proposals used images captured by a smartphone. In~\cite{ref:Lee2018ICCE}, superpixels were computed from grayscale images and the texture of the scene was analysed with wavelets. Then, a set of subtractions were applied to identify anomaly image patches that were marked as belonging to pothole category. Although the reported accuracy was high, there was a lack of thorough analysis over the dataset. In contrast to our evaluation on vehicle hardware platforms, their system was implemented on a high-end desktop computer. A similar hardware was used in \cite{ref:An2018ICCE} with the addition of a Nvidia GPU 770M to run different Deep Neural Networks (DNN) that classified images in two categories: one or more potholes in the image vs no potholes.  All models reported high accuracy above 96\%. However, the models were trained in a binary classification task without identifying the box around the pothole. Our DNN system does provide the detected box.

With regards to Deep Learning approaches, Maeda et. al.~\cite{ref:MaedaArxiv2018} has recently presented a road damage detection using DNNs. It defined 5 classes of road cracks and 3 mixed categories that gathered other types of hazards (rutting, bump, pothole, blurred markings). An image dataset with 15,435 annotations was built while driving in 7 cities of Japan. During our evaluation, we extracted images and labels from their dataset for comparison. However, their work was mainly focused on road cracks. A further review is provided in Section~\ref{sec:results}.

\subsection{Depth- and vision-based pothole detectors}\label{ssec:sotadepth}
The addition of depth data to the 2D visual clues is also present in the literature. \cite{ref:Moazzam2013ITSC} evaluated statically on the street the use of the Microsoft Kinect camera. However, this sensor is not well-suited for outdoors due to the InfraRed structured lighting that is mostly overridden under the presence of ambient IR from the sun.

Stereo vision for pothole detection was firstly introduced in~\cite{ref:Zhang2014ICASSP}. Three phases were proposed to construct the disparity map, fit a surface and apply a fixed height threshold to detect shallow road regions which potentially belonged to potholes. Stereo cameras were also employed in~\cite{ref:Mikhailiuk2016IST}, which implemented a real-time pothole detector on a digital signal processing unit based on a very similar algorithm. It must be noted that these approaches are sensitive to disparity estimation errors that would yield false pothole detections. In fact, ~\cite{ref:Mikhailiuk2016IST} required a stereo pair of cameras pointing downwards to the road. This is contrary to the tendency of placing cameras with XZ-plane nearly parallel to the ground such that other tasks can be done with the same images, e.g detection of objects/lanes ahead of the vehicle.

In~\cite{ref:KangICUFN2017} a 2D LiDAR and a camera were combined. Different pothole detection algorithms for each sensor modality were presented and the experiments were carried out in lab placing a shallow box to simulate the pothole. Thus, the setup was far from any real road scene. However, \cite{ref:SucgangIMECS2017} used real data and introduced the detection of speed bumps on LiDAR 3D data at the same time potholes were detected with basic image processing techniques.

To complete the literature review, there exist other private approaches without further technical details available: a Jaguar Land Rover warning system\cite{ref:JaguarPotholeDetection_2015}, a Google patent \cite{ref:GooglePatent_Potholes_2015} and a commercial black box camera for pothole detection~\cite{ref:Jo2015Sensors}. In the latter one, size, location and appearance of potholes were sent to a centralized server for off-line assessment and maintenance actions. The system was only tested on sunny weather.

Compared to above mentioned works, our proposal used an automotive grade camera installed on the front bumper of a car that can be deployed in new manufactured vehicles. Moreover, we applied latest progress in Deep Neural Networks (DNN) and tested the pothole detector in an embedded driverless vehicle platform. We built an image dataset with road scenes from different camera sources under varied environmental conditions to train the DNN models. Moreover, as part of the AUTOPILOT project~\cite{ref:AutopilotWebsite}, the detected potholes were reported in IoT-fashion to cloud services. The following section contextualizes our work within that project and 3 interrelated research areas.

\section{AUTOPILOT and related research areas}\label{sec:aviot}

In the last 30 years, there has been a huge progress in the field of Autonomous Vehicles (AV) in such extent that nowadays, there exist vehicles in the market that claim several autonomous driving functions and many more yet to come. From a computer vision perspective, \cite{ref:Janai17Arxiv} provides a wide literature review. 

Successful approaches in Artificial Intelligence (AI) have contributed to the progress. They have been proposed either as knowledge transfer from other research areas or as specific solutions for driver assistance an driverless vehicles.
Indeed, AI is a broad concept referring to the cognitive ability of machines as opposed to the natural intelligence of humans or other beings. In recent history, three terms have become very popular which represent different trends in the algorithmic implementation of AI. In chronological order: \emph{Pattern Recognition (PR)}~\cite{ref:Duda2010PR}, \emph{Machine Learning (ML)}~\cite{ref:Felzenszwalb2009TPAMI} and \emph{Deep Learning (DL)}~\cite{ref:Krizhevsky2012NIPS}. PR can be considered outdated but it belongs to the age when the premier international conference on Computer Vision and Pattern Recognition (CVPR) was born.
ML is still widely employed offering supervised and unsupervised learning methods. However, it has been recently outperformed by DL in many tasks and challenges at the cost of higher computational requirements and dedicated hardware.

In parallel, Internet of Things (IoT) extends the principle of internet services to include all possible types of devices, thus bringing the advance in accessibility of information from the virtual to the physical world. IoT can enable new capabilities in AV for various layers. From the communication for a given cooperation zone within vehicle range to the connection to cloud platforms with the aim of sharing information in larger zones, i.e. smart cities.

\begin{figure}[!h]
\centering
\includegraphics[width=0.85\columnwidth]{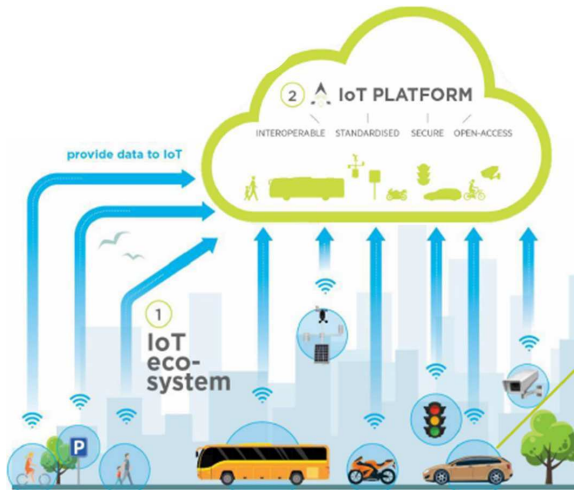}
 \caption{AUTOPILOT H2020 project illustration~\cite{ref:AutopilotWebsite}}
 \label{fig:autopilotleaflet}
\end{figure}

In 2016, the European Commission funded five Large-Scale Pilots (LSPs) on the IoT.  The \textbf{AUTOPILOT} H2020 project~\cite{ref:AutopilotWebsite} was selected as Pilot 5: \textit{autonomous vehicle in a connected environment}. AUTOPILOT concerns the use of IoT for enabling automated driving as it is illustrated in Fig.~\ref{fig:autopilotleaflet}. 

AUTOPILOT will deploy, test and demonstrate IoT-based automated services in five driving modes. The research work presented in this paper was applied to the Highway Pilot use case to locate and characterize road hazards like road surface anomalies, bumps, fallen objects, etc. 
The fusion of advances in AI, AV and IoT are studied to provide feasible solutions as the one proposed in this paper.

In AUTOPILOT, a cloud service merges the sensors' measurements from different IoT devices on board vehicles (camera, LiDAR, inertial and vibration sensors) and on the road side (mainly cameras). The goal is to provide incoming vehicles with meaningful warnings and driving recommendations to manage the hazards in a safer or more pleasant way. Moreover, the automated detection of the hazards enables the reporting to corresponding traffic or infrastructure authorities for maintenance or other monitoring and controlling actions.


Considering this background, our current research work has a direct impact on the operational performance of the new generation of ITS. One the one hand, recent progress in AI, more specifically Deep Learning, enables the integration of high-performing scene perception modules into vehicles equipped with multiple sensors. On the other hand, IoT enables the data collection from multiple sources to also implement AI solutions on the cloud. The following section describes the details of our pothole detector model based on latest AI.

\section{Learning pothole appearance with DNNs}\label{sec:learnpotholes}

\textbf{Definition of potholes:} Bowl-shaped holes of various sizes in the pavement surface that are greater than $175cm^2$ in area ($\sim 15cm$ diameter) (illustrative examples in Fig.~\ref{fig:potholetraincollage}).

A set of factors generate road distress that leads to the formation of potholes~\cite{ref:NienaberSSCI2015}.
The most relevant causes are the volume of traffic, the axle load of heavy vehicles (buses and trucks) and the environmental conditions (day/night temperature contrast, snow, rain). Moreover, the type of road surface material, the underlying terrain and construction techniques influence the quality and resistance of the pavement.

Typically, the first phase is the formation of cracks that allow water to seep through. Next, when vehicles drive over them, the water is pushed in many directions around the initial cracks. Eventually, cavities will appear in the asphalt growing in size until they reach pothole dimensions. Besides, the wider the cavity the quicker the process. Therefore, automated detection and notification of potholes helps preventing progressive road deterioration. Our research work proposes to learn the visual appearance of potholes with Deep Neural Networks.


\subsection{Visual appearance of potholes}\label{ssec:vispotholes}
\begin{figure}[!h]
\centering
\includegraphics[width=0.9\columnwidth]{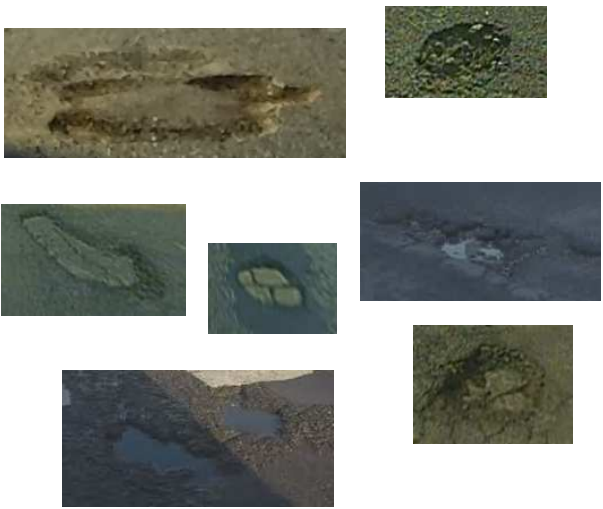}
 \caption{Samples of annotated potholes. They come from several sources and represent varied places, environmental conditions and camera viewpoints.}
 \label{fig:potholetraincollage}
\end{figure}

We have collected images with several types of potholes under different illumination and weather conditions as depicted in Fig.~\ref{fig:potholetraincollage}. The most common potholes show a pronounced edge describing an elliptical shape. However, there are some that describe a more square-like shape and some others that do not have a pronounced edge. They might appear darker or brighter on the background pavement. The deepest ones look darker because of the shadow of the edge, while in the flat ones it is possible to see the ground or gravel inside the pothole. Besides, some potholes appear filled with water and might also reflect surrounding scene in the surface.

Attending to this intra-class variance, pothole detection based on classic computer vision has limitations. The literature reviewed in Section~\ref{ssec:sotavision} presented ad-hoc image processing filters and heuristics, which were biased towards specific datasets and conditions. We propose to take successful DNN object detection models as a base, thus to automatically learn more discriminative visual features of the potholes in varied scenes.

\subsection{Deep Learning for the detection of potholes}\label{ssec:dnnobjdet}

Object detection approaches based on engineered features and Machine Learning classifiers have been very fruitful until recent times~\cite{ref:Felzenszwalb2009TPAMI}. However, when applied to different tasks or adapted for additional challenges they required intensive parameter tweaking and dimensionality reduction~\cite{ref:YebesSensors2015}. 

Recent trends on Deep Learning have achieved impressive detection performance on various tasks~\cite{ref:Krizhevsky2012NIPS} including object detection~\cite{ref:HuangArxiv2017}. Moreover, several DNN based models are publicly shared and can be used as initialization step for further training and fine-tunning in different object classification tasks~\cite{ref:TensorFlow_ModelZoo}. We make use of this progress in AI for the detection of potholes.

Based on the investigation in~\cite{ref:HuangArxiv2017} and the entries reported in the TensorFlow model Zoo~\cite{ref:TensorFlow_ModelZoo}, we selected 4 models that have shown high detection ratios at reasonable processing costs. Attending to detection performance we chose 3 different configurations of the architecture \emph{Faster R-CNN}. Besides, we evaluated the Single Shot multibox Detector (\emph{SSD}) because it is targeted for mobile applications. They are explained in the next paragraphs.\\

\begin{figure}[htb]
    \centering
    \includegraphics[width=0.6\columnwidth]{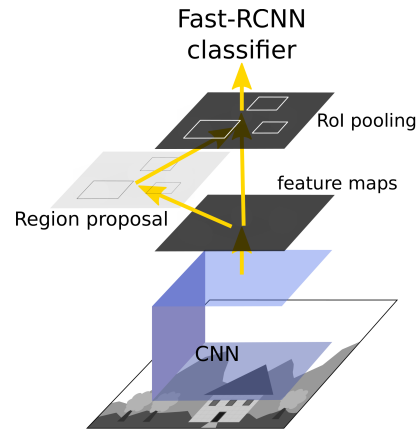}
    \caption{Simplified diagram of Faster R-CNN for illustration purposes}
    \label{fig:fasterRcnnDiag}
\end{figure}

The Faster R-CNN architecture uses a system called Region-based Convolutional Neural Network (R-CNN). As opposed to a brute-force sliding window approach over image space, regions of interest are proposed and warped to fixed size and then, they are individually fed into a CNN for classification and bounding box refinement.
The Faster R-CNN network architecture consists of three major blocks (see Fig.\ref{fig:fasterRcnnDiag}):
\begin{enumerate}
\item A CNN to extract features. In brief, the weights of several convolutional filters are learned in order to calculate the feature maps.
\item A Region Proposal Network (RPN) that acts as attention model. It proposes regions that may contain an object. During training, it learns to determine whether a certain region has an object or not and the appropriate shapes and sizes. This is the multi-task loss:
\begin{multline}
    L(\{p_i\},\{t_i\})=\frac{1}{N_{cls}}\sum_i L_{cls}(p_i,p_i^*) + \\
      + \lambda\frac{1}{N_{reg}}\sum_i p_i^* L_{reg}(\boldsymbol{t}_i,\boldsymbol{t}_i^*)
    \label{eq:fasterrcnnmultiloss}
\end{multline}

where $i$ is the ith candidate region (called anchor in \cite{ref:Ren2017TPAMIFasterRCNN}), $p_i$ is the predicted probability to be an object, $p_i^*$ is 1 if the candidate truly contains an object and 0 otherwise. $\boldsymbol{t}_i$ are the coordinates of the candidate region, $\boldsymbol{t}_i^*$ is the ground-truth box, $L_{cls}$ is the log loss over two classes (object/non object) and $L_{reg}$ the regression defined as the robust smooth L1 function\cite{ref:Ren2017TPAMIFasterRCNN}.

\item The last block is the object classifier Fast R-CNN~\cite{ref:Girshick2015CVPRFastRCNN} that receives the proposed regions, assigns a class to them and it also refines the bounding boxes using a regressor. The loss function in training is very similar to the one in the RPN.
\end{enumerate} 

We have fine-tuned these blocks end-to-end using stochastic gradient descent optimization with back propagation.

The seminal method was very slow to train because of the computation of features for every region. Then it eventually evolved into Faster R-CNN~\cite{ref:Ren2017TPAMIFasterRCNN}. Despite its "Faster" prefix, it is slower than SSD because it re-runs the cropped regions through the feature extractor. In fact, we selected three Faster R-CNN models that differ in the feature extractor because the choice of feature extractor has a higher impact in Faster R-CNN models than in SSD. The choices are introduced below:

\noindent\textbf{Faster R-CNN Inception v2}. The feature extractor in this approach provides batch normalization for accelerating the model training and it has yielded high accuracy~\cite{ref:HowardArxiv2017Mobilenets}. Compared to v1, it is more efficient by factorizing the convolution operations. Also, it is a wider network to avoid losing visual details that may happen in v1 which is deeper as road potholes typically represent small portions of the images.

\noindent\textbf{Faster R-CNN Resnet101}. This model training uses Resnet101 that stands for Residual Network with 101 layers~\cite{ref:HowardArxiv2017Mobilenets} and has achieved high success on many competitions. This type of networks try to learn residuals which are short-cut connections between layers. The approach allows to train deeper models without degradation.

\noindent\textbf{Faster R-CNN Inception-Resnet v2 (atrous)}. This third model uses a hybrid feature extractor that combines Inception and Resnet, second version (v2), yielding improved recognition performance as reported in~\cite{ref:Szegedy2017Arxiv}. Moreover, the "a trous" (\emph{with holes in}) option employs dilated convolutions, which provide a wider field of view at the same computational cost towards achieving better accuracies.

Opposed to above ones, the Single Shot multibox Detector (SSD) uses a single feed-forward convolutional network~\cite{ref:Liu2015ECCVSSD}. It is a faster and simpler method because it discards the proposal generation phase and feature re-sampling. It uses a set of default boxes with different aspect ratios and scales per each feature map location. These boxes are equivalent to anchors in Faster R-CNN and represent priors manually chosen as described in~\cite{ref:Liu2015ECCVSSD}. For the sake of clarity, the training loss is  reproduced below. It is a combination of two functions that capture object confidence and location error:
\begin{equation}
    L(x,c,l,g)=\frac{1}{N}(L_{conf}(x,c) + L_{loc}(x,l,g) ),
    \label{eq:ssdloss}
\end{equation}

where N is the number of matched default boxes and $x$ is 1 to mark detections correctly matched to ground-truth or zero otherwise. $L_{conf}$ is the softmax loss over classes confidences ($c$) and the localization loss $L_{loc}$ is a smooth L1 function between predicted ($l$) and ground-truth box ($g$) parameters.

\noindent\textbf{SSD Mobilenet v2}.  We selected \emph{Mobilenet version 2} as feature extraction network~\cite{ref:HowardArxiv2017Mobilenets}, which has been specifically designed for mobile vision applications with limited resources. It uses depth-wise separable convolutions and the version2 incorporates a set of improvements. One of them are residual connections as in ResNet but applying thinner bottleneck layers~\cite{ref:Mobilenetv22018} which makes it more efficient than v1.\\

\subsection{Fine-tune DNN models for the detection of road potholes}\label{ssec:potholednns}

The selected models were successfully evaluated on multi-object detection challenges and we fine-tuned them for a single class: road potholes. We downloaded the above mentioned pre-trained DNN models~\cite{ref:TensorFlow_ModelZoo}, which are used as initialization. The weights of the feature extractor layers are automatically fine-tuned during the training. The weights of the remaining layers that perform the classification and object localization tasks, are randomly initialized and learned towards the visual detection of potholes. This is a established practice when applying Deep Learning models. As opposed to hand-engineered features such as SIFT, SURF, HOG which are well-known in computer vision literature, these DNN models use automatically learned features from the database \emph{Common Objects in Context (COCO)}~\cite{ref:COCO}. This large-scale object detection dataset contains more than 200K images distributed in 90 object categories depicting real world scenes. It is sufficiently large to learn rich visual features of naturalistic scenes that are encoded in the weights of the neural networks. These features cannot be learned using a limited number of object labels, like our pothole case. Hence, the pre-trained models transfer the knowledge in order to train the pothole detector.

For the fine-tuning of the models, we modified several parameters after conducting an analysis on the dataset. In the following experimental section, we first described the dataset and then, the details of the changed parameters in Section~\ref{ssec:experiments_potholes}.

\section{Experimental section}\label{sec:results}

\subsection{Dataset}\label{ssec:datasets}

The images used in our research have been obtained from varied sources and belong to different places in the world.  Only a subset was already labelled and several images have been manually annotated with boxes around potholes. Our motivation to build this dataset has been twofold: I) gathering enough data samples for fine tunning and evaluation and II) increase variance in pothole appearance, camera viewpoints and road scenes. We have divided the images in training, validation and test sets.

\subsubsection{\textbf{Training and validation sets}}\label{ssec:datatrain}
The database consists of 5,874 images from which 5,774 have been used for fine-tunning and 100 have been randomly picked for validation. In the training set there is a total amount of 9,716 potholes while in the validation set the number of potholes is 171. All the images have been captured from a vehicle, with different camera placements and viewpoints. We constructed the training set from these sources:
\begin{itemize}
\item 4,030 images of size 3680$\times$2760 pixels and their labels were obtained from~\cite{ref:NienaberSUN2015}. The images were captured as regular snapshots from a GoPro camera attached to the inner side of a vehicle windscreen.
\item 1,644 images have been sampled from AUTOPILOT videos of VALEO in the surroundings of Paris and potholes have been manually labelled. These images have a resolution of 1280$\times$800 pixels. The sequences were captured by an automotive-grade fish-eye camera mounted on the front bumper of a Volkswagen Tiguan 2. We removed the radial distortion as pre-processing step using the calibration parameters of the camera. Then, the undistorted images were added to the dataset.
\item The remaining 100 images (up to 5,874) have been captured from the Google Earth Pro street-view tool at a resolution of 1236$\times$804 pixels. They have been also manually labelled. Most of these images containing potholes belong to streets of Tirana in Albania and some have been found in San Sebastian and Bilbao in Spain.
\end{itemize}   

\begin{figure}[!ht]
\centering
\subfloat[South Africa~\cite{ref:NienaberSUN2015}]{\includegraphics[width=0.48\columnwidth]{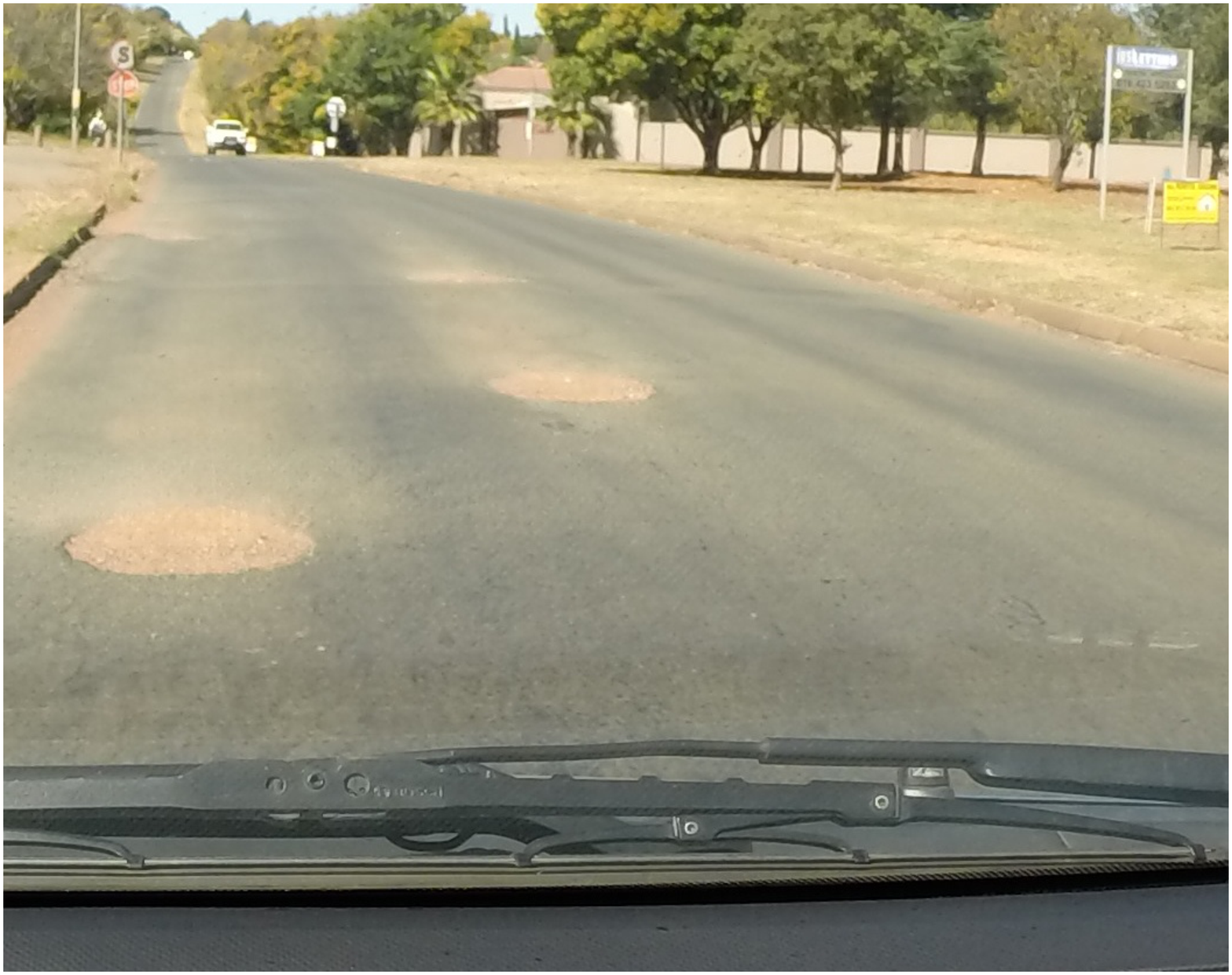}}%
\hfill
\subfloat[Google - San Sebasti{\'{a}}n]{\includegraphics[width=0.48\columnwidth]{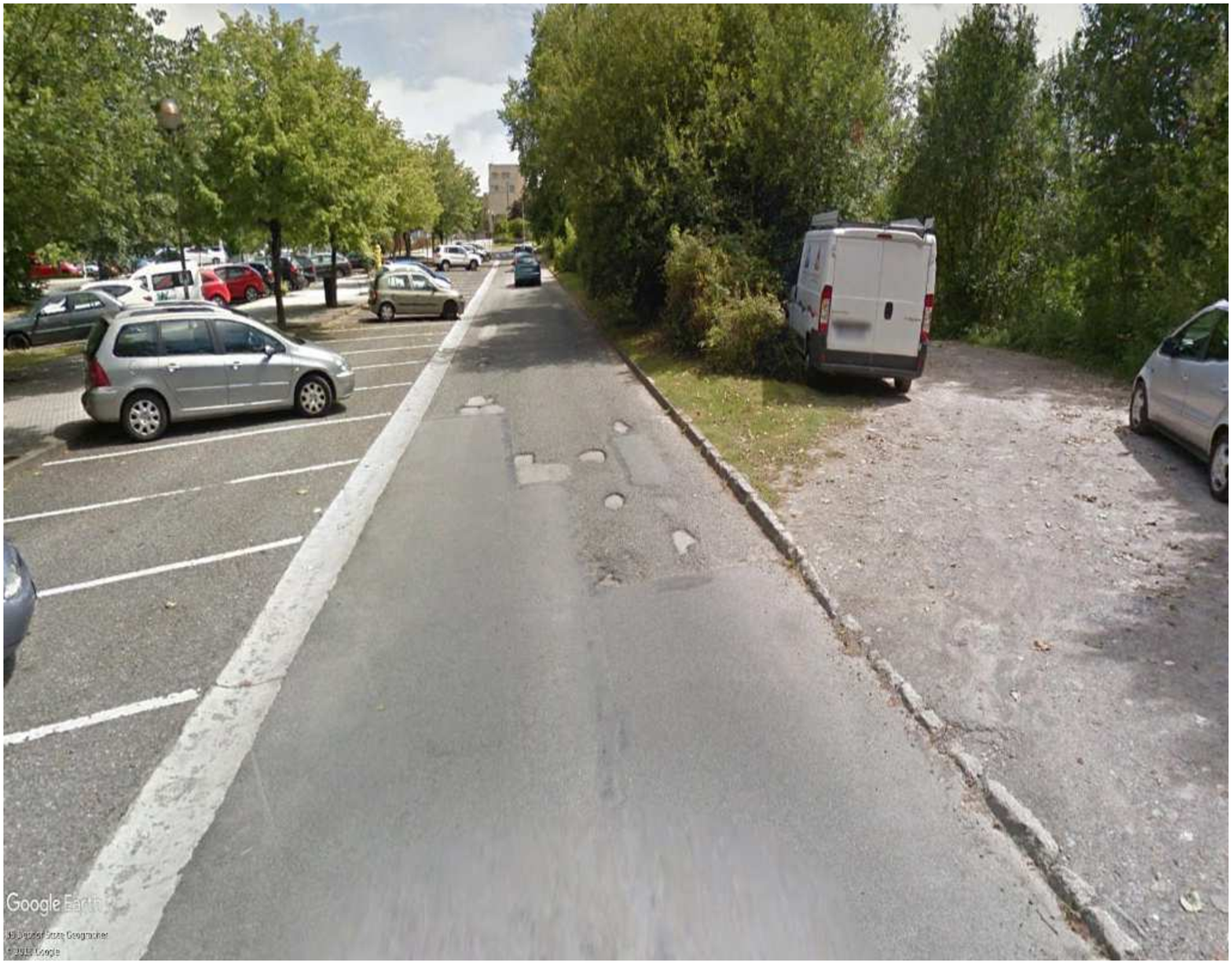}}\\
\subfloat[Valeo - Paris]{\includegraphics[width=0.48\columnwidth]{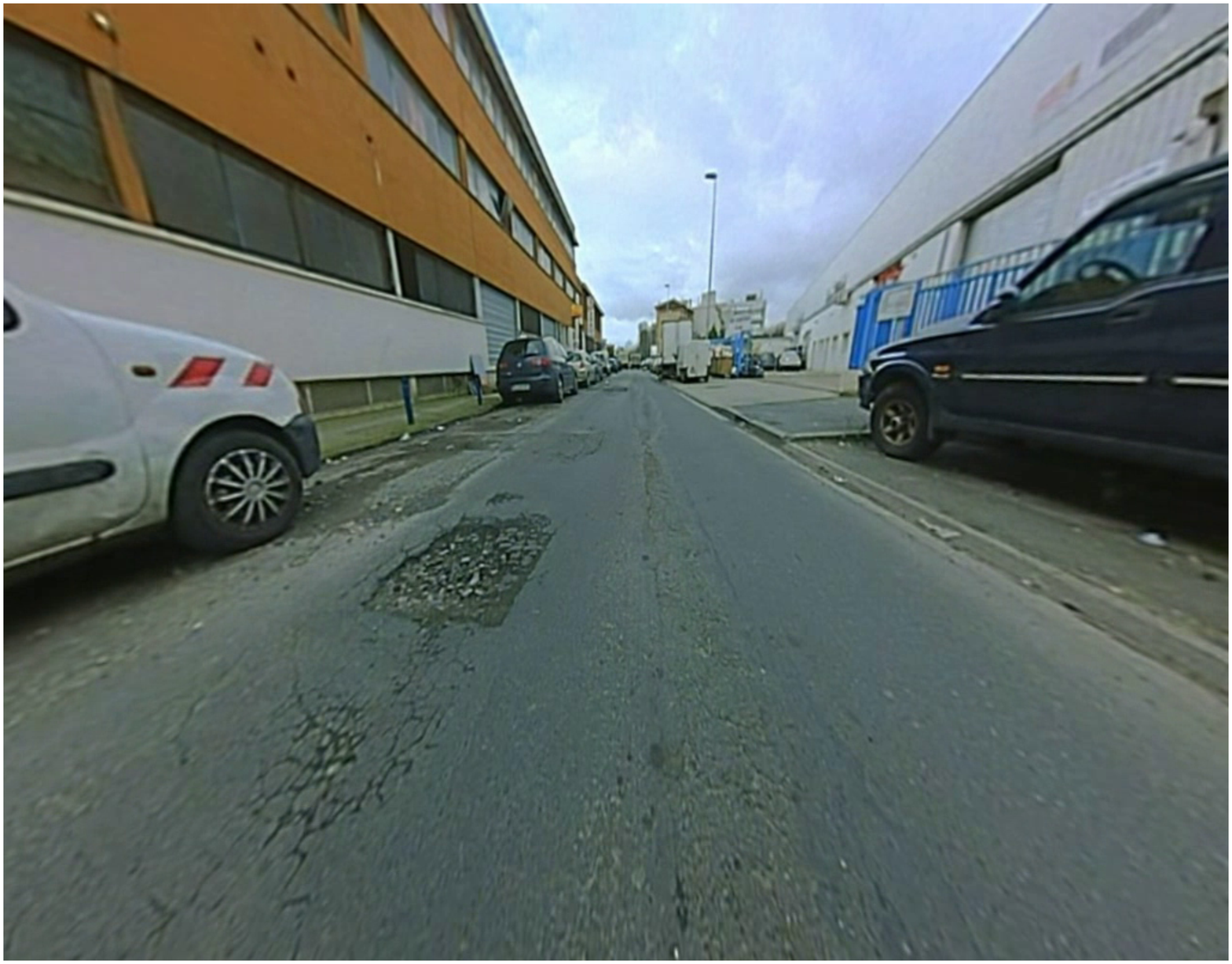}}%
\hfill
\subfloat[Google - Tirana]{\includegraphics[width=0.48\columnwidth]{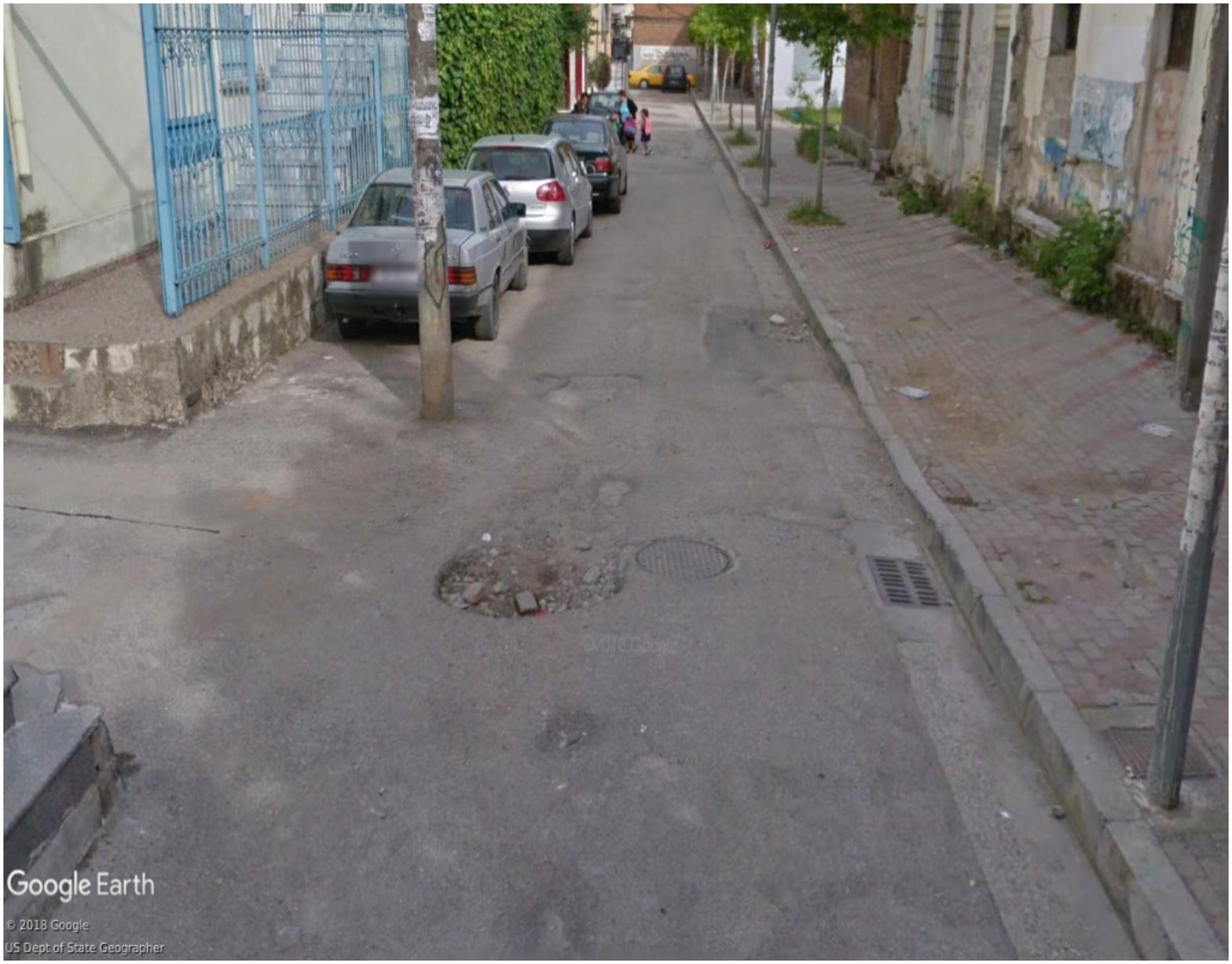}}%
\caption{Sample images of the constructed training+validation image set.}
\label{fig:trainimgs}
\end{figure}

We used the tool \emph{Label Img}~\cite{ref:TzutalinLabelImg} to annotate the boxes enclosing road potholes. The images were slightly cropped near the borders to remove some scene background and resized to 1024$\times$800 pixels (see Fig.~\ref{fig:trainimgs}). The motivation for this size is explained in Section~\ref{ssec:experiments_potholes}.

\begin{figure}[!ht]
    \centering
    \scalebox{0.6}{\fontsize{12}{5}\selectfont\input{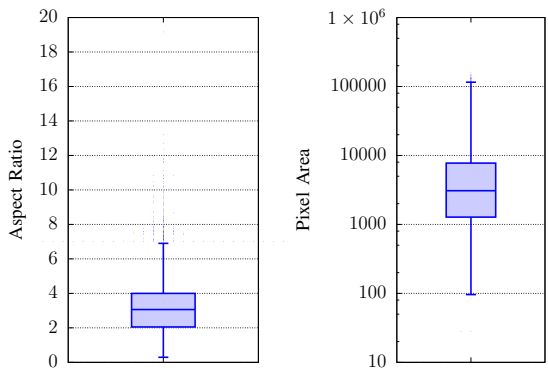}}
    \caption{Distribution of the aspect ratio and area in pixels of the annotated potholes for the train dataset. The boxplot on the left has a linear representation of the aspect ratios. The quartiles are $Q_1=2.053, median=3.062, Q_3=4.0$. The boxplot on the right shows the area in pixels and has a logarithmic axis for the purpose of visualization. The quartiles are $Q_1=1276, median=3081, Q_3=7744$ pixels }
    \label{fig:boxplot_area}
\end{figure}

We analysed the distributions of the aspect ratio and area of the annotated potholes for fine-tunning the training parameters of the DNNs. We provide box plots in Fig.~\ref{fig:boxplot_area}. The size and location of the blue box indicates those values from the first (25\%) to the third quartiles (75\%) and the interquartile distance ($IQR=Q_{3}-Q_{1}$). The line inside the box represents the median value. The whiskers represent the highest observation below the upper limit ($L_{upper}=Q_{3}+1.5 \cdot IQR$), and the lowest observation above the lower limit ($L_{lower}=Q_{1}-1.5\cdot IQR$). The points outside those two limits are considered outliers.

\subsubsection{\textbf{Test set}}\label{ssec:datatest}
The test images have been collected from different sources as well. In first place, some additional images have been fetched from the sequences recorded in AUTOPILOT project with VALEO vehicle. The videos were captured in different dates and environmental conditions compared to the training set. The routes were also nearby Paris sharing some streets with the training sequences. Indeed, we manually selected 265 images with 128 potholes in challenging scenarios, which included dark lighting (tunnels), rough weather conditions, traffic jams and roads with bumps, stains and manholes. The latter ones being potential cases of false positive detections as potholes. 

The second source is a youtube video\footnote{\url{https://www.youtube.com/watch?v=BQo87tGRM74}} recorded from a dash-cam in a car driving through a road stretch in the area of Willamette National Forest, Oregon, US. We extracted 482 frames and manually labelled 475 potholes.

Additionally, we re-annotated some of the images in~\cite{ref:MaedaArxiv2018}. The road damage inspection presented in that paper released a public dataset with 9,892 pictures captured with smartphones. Among the  classes of hazards identified in~\cite{ref:MaedaArxiv2018}, one was categorized as \emph{Rutting, bump, pothole and separation}. It is a very broad class with loose large boxes around the hazards. Hence, we reviewed them to generate new labels only for potholes discarding the remaining annotations. As a result, 62 potholes have been selected within 54 images.

\subsection{Experimental details of DNN models of road potholes}\label{ssec:experiments_potholes}
As introduced in Section~\ref{ssec:potholednns}, we fine-tuned 4 DNN models for the detection of road potholes. In first place, we studied the nature of the dataset and we fixed the size of the input layer to 1024$\times$800 pixels. We found that the median area of potholes corresponded to a 0.37\% of the original image size. Given a reduced resolution of 600$\times$600 as reported in~\cite{ref:TensorFlow_ModelZoo}, $1350 \simeq 36^2$ pixels is the average area of the potholes in the images. Besides, the 1:1 aspect ratio involves warping, cropping and resizing the original images. Consequently, road areas on the left and right sides in front of the car are cropped out and the visual appearance of the scene and the potholes are deformed and reduced in granularity. Therefore, we decided to use a larger window size with an aspect ratio similar to the original resolution of the images, achieving a median area of $3081$ pixels among the annotated potholes.

In addition, we applied a set of adjustments for fine-tuning the 3 Faster R-CNN models.
\begin{itemize}
\item Added the aspect ratios of 1:3 and 1:4 after analysing the dataset (see Fig.~\ref{fig:boxplot_area}). Thus, the aspect ratios during learning were $[.5, 1, 2, 3, 4]$. 
\item Reduced the maximum number of region proposals for Faster R-CNN models from 300 to 100, with the aim of decreasing detection time without losing performance~\cite{ref:HuangArxiv2017}.
\item Due to the limited size of our pothole database, we increased the number of training samples by data augmentation. Among several options for fine-tunning Faster R-CNN models, we selected \emph{random\_adjust\_brightness} and \emph{random\_horizontal\_flip} because of their effectiveness in previous research experience for object detection.
\item We enabled the drop-out feature in the second stage of the Faster R-CNN training to prevent over-fitting. Basically, this option randomly drop units and their connections from the network, which  prevents the units from co-adapting too much.
\item The number of steps employed for fine-tuning the 3 models was 2 millions. We picked this number observing the performance of the trained models on the validation subset. The loss function was stable without showing too much improvement, thus we decided to stop at 2M steps, which corresponds to 173 epochs.
\end{itemize}

Similarly, we applied the same adjustments for fine-tuning the \emph{SSD Mobilenet v2} model but using a a squared size of 800 $\times$ 800 pixels.

\subsection{Implementation details for training \& evaluation}\label{ssec:implement}

We include in this section the relevant hardware and software details of the experiments. For the fine-tuning we have used a barebone PC equipped with 2 $\times$ Xeon 20-Core $@$ 2.2GHz, 16 $\times$ 32GB RAM modules and 8 $\times$ GPGPU Nvidia Tesla P100M with 16GB of memory each one. It must be noted that only one GPGPU was used during the learning of a given DNN model. The software environment included Ubuntu OS, Tensorflow 1.7 with Tensorboard for the visualization and validation of the learning process~\cite{ref:tensorflowWP2015}.
      
For the evaluation of the models, we used the same PC and also the Nvidia DrivePX2 Autochauffeur\footnote{https://www.nvidia.com/en-us/self-driving-cars/drive-platform/}, which is specifically designed for research and embedding AI solutions in driverless vehicles.  This platform has two Nvidia Tegra X2 SoCs (known as TegraA and TegraB), where each SoC contains 2 Denver cores, 4 ARM A57 cores and a Pascal GPU. This replication is for redundancy purposes and not for parallelization. The dGPU has 1152 CUDA cores and 4GB of memory while the iGPU has 256 cores and uses the shared system memory 7GB. Besides, the device includes a set of interfaces (GMSL, USB, Ethernet, CAN, FlexRay, etc.) for different sensors like cameras, LiDAR, GPS, CAN bus signals, etc.
The OS is based on Ubuntu Linux distribution and we carried out the tests with version \emph{Drive 5.0.5.0a}, which includes CUDA9. Additionally, Tensorflow 1.7 was installed to load the learned DNN models and automatically find potholes in road scene images.

\subsection{Results}\label{ssec:results}

In our research, we considered the \emph{mean Average Precision} (mAP) as detection performance metric and the higher the mAP, the better. We analysed two different approaches to obtain it: COCO and PASCAL.  Both of them count the number of true/false positives and false negatives to calculate precision and recall. Next, the Average Precision is obtained as 
\begin{equation}
   AP = \frac{1}{11} \sum_{r \in \{0, 0.1, ... , 1\}} \underset{\tilde{r}:\tilde{r}\geq r}{max} \; p(\tilde{r}) ,
   \label{eq:AP}
\end{equation}
where $r$ and $p$ are the recall and precision values respectively and 11 equally spaced sampled recall points are used. 

To determine object localization accuracy, the Intersection over Union (IoU) measures the overlap between predicted and ground-truth bounding boxes around objects. For the estimation of mAP, COCO defines the \emph{mean Average Precision} over all object classes and multiple values of IoU: precision-recall curves are evaluated at $IoU = [.5:.95]$ in steps of 0.05. In PASCAL mAP, only one value of IoU=0.5 is reported. Then, using COCO strategy, models showing higher precision in object localization are rewarded. For our analysis, we have also decided to report PASCAL mAP $@$ IoU=0.4. Precisely annotating the location and boxes around potholes is challenging, mostly on road scenes with deteriorated surfaces or road defects that could be labelled as one or several potholes. Observing the results, we realized that many pothole detections on the road surface could be considered as good by a human viewer but they were classified as false positives or false negatives by the evaluation algorithm. Therefore, with the aim of reducing the number of FN and FP we also report the mAP $@$ IoU=0.4.

Before presenting our results, for reference, we reproduced here the mAP and processing speed of the pre-trained models as reported in COCO entries~\cite{ref:TensorFlow_ModelZoo}. The timings were obtained for images of 600$\times$600 pixels and running on a Nvidia GeForce GTX TITAN X. 
\begin{enumerate}
\item ssd\_mobilenet\_v2\_coco ($t=31ms$, $mAP=22$)
\item faster\_rcnn\_inception\_v2\_coco ($t=58ms$, $mAP=28$)
\item faster\_rcnn\_resnet101\_coco ($t=106ms$, $mAP=32$)
\item faster\_rcnn\_inception\_resnet\_v2\_atrous\_coco ($t=620ms$, $mAP=37$)
\end{enumerate}

Running time and detection performance are correlated: the higher mAP, the higher processing delay. The reason is due to neural network architecture complexity and number of layers, provided that the size of the input layer is constant for all of them. A thorough analysis can be found in~\cite{ref:HuangArxiv2017}. 

Next, we report the results obtained after fine-tuning \emph{SSD Mobilenet v2} for potholes and evaluating the model on the test set. The AP was 33.62\% and 45.57\% $@$ IoU 0.5 and 0.4, respectively and took approximately 455ms per image when executed in Nvidia DrivePX2. These values show a very low detection performance despite the  fine-tunning of some parameters. We investigated the main reasons and reached the conclusion that in SSD Mobilenet the discretization in bounding boxes and their separate analysis does not account for neighbouring pixels, which provide useful context information. Despite the initial expectations  of this research to implement a mobile embedded Deep Neural Network, the SSD Mobilenet is not well suited to detect objects that strongly depend on the appearance of surrounding scene, i.e. potholes vs road appearance. In addition, by design it does not cope well with small bounding boxes compared to the image size. However, our dataset contains small annotated boxes compared to the size of the image. 

\begin{figure}[!h]
    \centering
    \scalebox{0.7}{\fontsize{10}{5}\selectfont\input{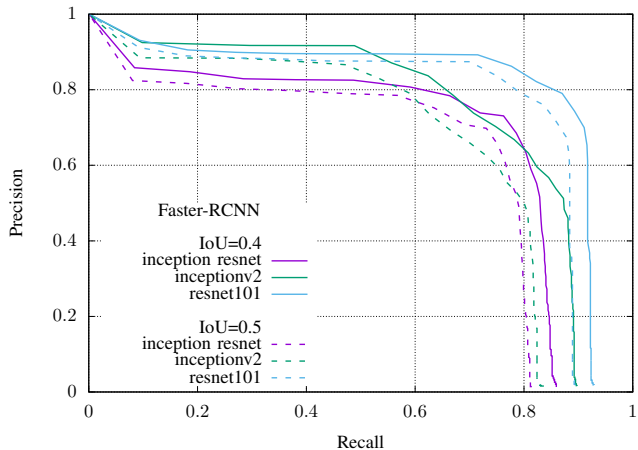}}
    \caption{Precision-recall curves for the test set. Three Faster R-CNN models with a different feature extractor network and two values of IoU are compared. Due to the nature of road potholes, there is a larger error in the localization accuracy of the ground-truth bounding boxes. Thus, we opted to report IoU = 0.4 in addition to the commonly used 0.5 value. The aim is reducing false negatives and false positives in road scenes with deteriorated surfaces where manual pothole labelling is also challenging.}
    \label{fig:plots_prtest}
\end{figure}

The remaining of this section is dedicated to the results of the selected Faster R-CNN models. Figure~\ref{fig:plots_prtest} presents the precision-recall curves on the test set for the 3 Faster R-CNN models at two different values of IoU and Table~\ref{table:PrecRecvalues} shows precision and recall based on PASCAL evaluation metrics. Table~\ref{table:mAPvalues} shows the mean Average Precision for each of them. Moreover, running times are shown in Table~\ref{table:inftimes}. The model names have been shortened\footnote{For visualization purposes the prefix "Faster R-CNN" has been omitted\label{fn:table}}.
\begin{table}[!ht]
\centering
\caption{Precision (p) and recall (r) of the evaluated pothole detectors based on Faster R-CNN models.\textsuperscript{\ref{fn:table}}}
\label{table:PrecRecvalues}
\renewcommand{\arraystretch}{1.5}
\begin{tabular}{ll|c|c|}
\hline
\multicolumn{1}{|l|}{\textbf{IoU}}         & \multicolumn{1}{c|}{\textbf{Model}} & \multicolumn{1}{l|}{\textbf{p (\%)}} & \multicolumn{1}{l|}{\textbf{r (\%)}} \\ \hline
\multicolumn{1}{|c|}{\multirow{3}{*}{0.5}} & inception\_v2                       & 70.10     & 65.85 \\
\cline{2-4} 
\multicolumn{1}{|c|}{}                     & resnet101                 & \textbf{83.48}    & \textbf{75.23}\\ 
\cline{2-4} 
\multicolumn{1}{|c|}{}                     & inception\_resnet\_v2\_atrous       & 73.08   & 69.78  \\ 
\hline \hline
\multicolumn{1}{|l|}{\multirow{3}{*}{0.4}} & inception\_v2                       & 74.7    & 70.26  \\ 
\cline{2-4} 
\multicolumn{1}{|l|}{}                     & resnet101                 & \textbf{82.27}   & \textbf{82.16}  \\ 
\cline{2-4} 
\multicolumn{1}{|l|}{}                     & inception\_resnet\_v2\_atrous       & 76.5        & 73.07  \\ 
\hline
\end{tabular}
\end{table}
\begin{table}[!ht]
\centering
\caption{Mean average precision of the 3 pothole detectors.\textsuperscript{\ref{fn:table}}}
\label{table:mAPvalues}
\renewcommand{\arraystretch}{1.5}
\begin{tabular}{ll|c|c|}
\cline{3-4}
                                           &                                     & \multicolumn{2}{c|}{\textbf{mAP(\textasciicircum{}1)}}                    \\ \hline
\multicolumn{1}{|l|}{\textbf{IoU}}         & \multicolumn{1}{c|}{\textbf{Model}} & \multicolumn{1}{l|}{\textbf{PASCAL}} & \multicolumn{1}{l|}{\textbf{COCO}} \\ \hline
\multicolumn{1}{|c|}{\multirow{3}{*}{0.5}} & inception\_v2                       & 67.05          & 27.51 \\
\cline{2-4} 
\multicolumn{1}{|c|}{}                     & resnet101                           & \textbf{77.49} & \textbf{31.12}\\ 
\cline{2-4} 
\multicolumn{1}{|c|}{}                     & inception\_resnet\_v2\_atrous       &  68.71         & 26.67 \\ 
\hline \hline
\multicolumn{1}{|l|}{\multirow{3}{*}{0.4}} & inception\_v2                       &  75.45         & ---  \\ 
\cline{2-4} 
\multicolumn{1}{|l|}{}                     & resnet101                           & \textbf{82.02} & --- \\ 
\cline{2-4} 
\multicolumn{1}{|l|}{}                     & inception\_resnet\_v2\_atrous       & 69.72          & --- \\ 
\hline
\end{tabular}
\end{table}

\begin{table}[ht]
\centering
\caption{Average inference time for the 3 pothole detectors \textsuperscript{\ref{fn:table}} and different gpgpu devices. }
\label{table:inftimes}
\renewcommand{\arraystretch}{1.5}
\begin{tabular}{l|c|c|}
\cline{2-3}
                                                    & \multicolumn{2}{c|}{\textbf{Inference time (ms)}}                                                \\ \hline
\multicolumn{1}{|l|}{\textbf{Models}}               & \multicolumn{1}{l|}{\textbf{Tesla P100M 16GB}} & \multicolumn{1}{l|}{\textbf{DrivePX2 TegraA}} \\ \hline
\multicolumn{1}{|l|}{inception\_v2}                 & 53.2                                           & 177.1                                           \\ \hline
\multicolumn{1}{|l|}{resnet101}                     & 94.2                                           & 432.7                                           \\ \hline
\multicolumn{1}{|l|}{inception\_resnet\_v2\_atrous} & 172.1                                          & 732.9                                           \\ \hline
\end{tabular}
\end{table}

As it can be observed from the resulting values, \textbf{Faster R-CNN Resnet101} yields the highest detection performance both in PASCAL and COCO evaluation metrics. It is not the slowest DNN but it requires on average 432.7ms per frame on the Nvidia DrivePX2. For the goals of detecting potholes and reporting them to an IoT platform (AUTOPILOT project), it is a valid time. There are not real-time requirements because detected potholes will be monitored from control centres and broadcasted to warn other road users when sufficient confidence is reached after several repeated detections on a given map location. For vehicle reactive manoeuvres upon detection, further research is needed to optimize the neural networks without losing too much performance in order to obtain processing gains. A significant reduction of size on the input layer could importantly increase the number of false detections due to the small size of potholes compared to the rest of the image. Thus, other DNN optimization strategies should be explored, like enlarging the dataset, retraining on mobile or pruned networks from scratch and transforming DNNs with software optimization tools for its execution on specific hardware. 

Also, from the results, it is surprising that \emph{Faster R-CNN Inception Resnet v2 (a trous)} obtains lower AP, which is contrary to expected performance. Indeed, comparing the AP $@$ IoU 0.5 and 0.4 it can be seen that a slight gain of 1\% is achieved, meaning that this DNN obtains more false negatives and more false positives that do not overlap correctly to ground-truth potholes. Besides, the validation p-r curve during fine-tuning shows a lower performance than the other models. On the one hand, the complexity of this network and the limited number of training samples can cause weak propagation of relevant signals for the class potholes. On the other hand, the "a trous" option might be negatively affecting the selection of features in the surroundings on the potholes. Also, we suspect that some over-fitting is also happening due to a larger gap between validation and testing performance compared to the other Faster R-CNN approaches.

Finally, \emph{Faster R-CNN Inception v2} yields modest values of AP at lower processing cost, which is also a valid pothole detector for the aim of Highway Pilot use case within AUTOPILOT. Rates up to 6fps when running on NVidia DrivePX2 could be achieved. 

\begin{figure*}[!t]
\centering
\includegraphics[width=0.24\textwidth]{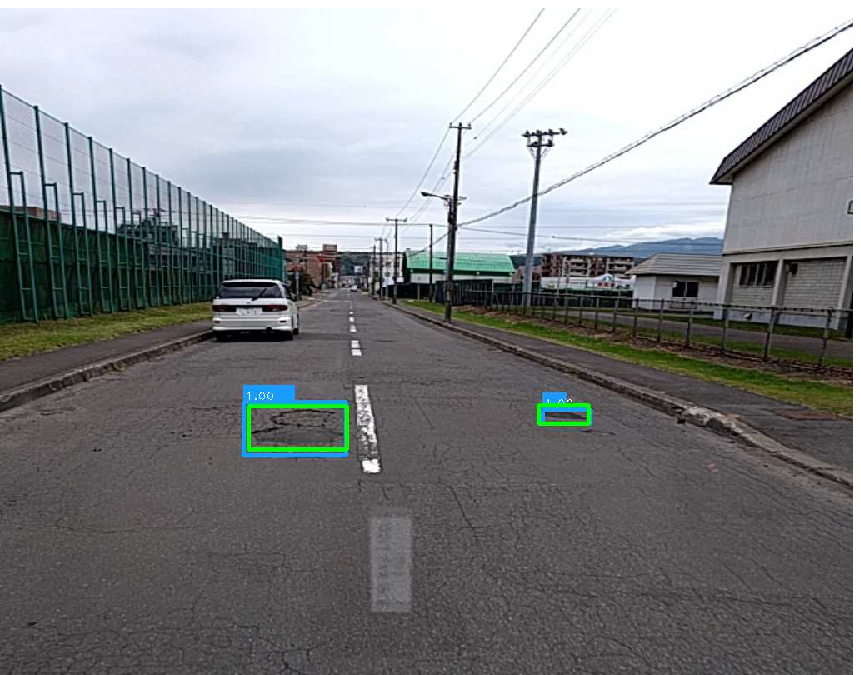}%
\hfil
\includegraphics[width=0.24\textwidth]{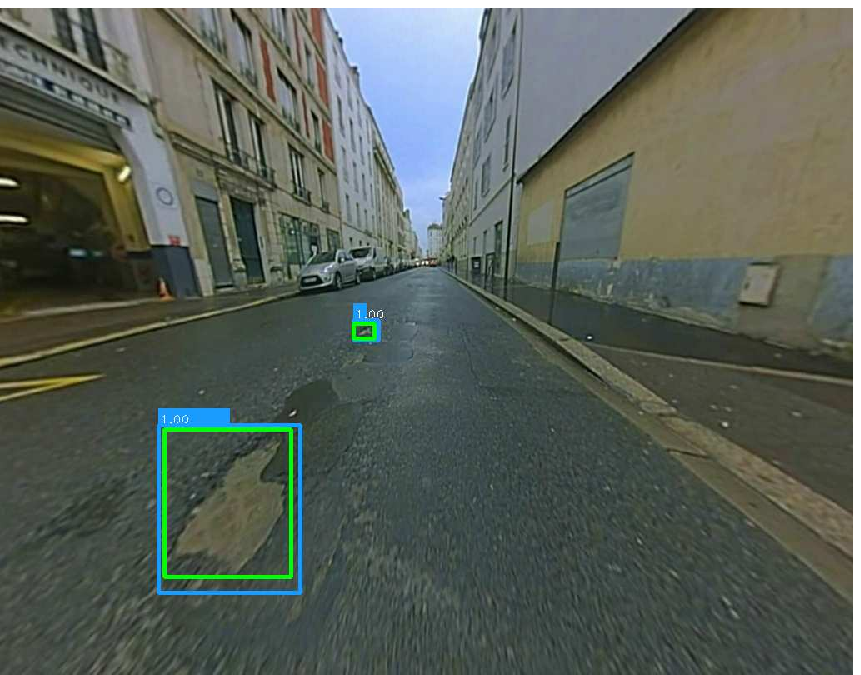}
\hfil
\includegraphics[width=0.24\textwidth]{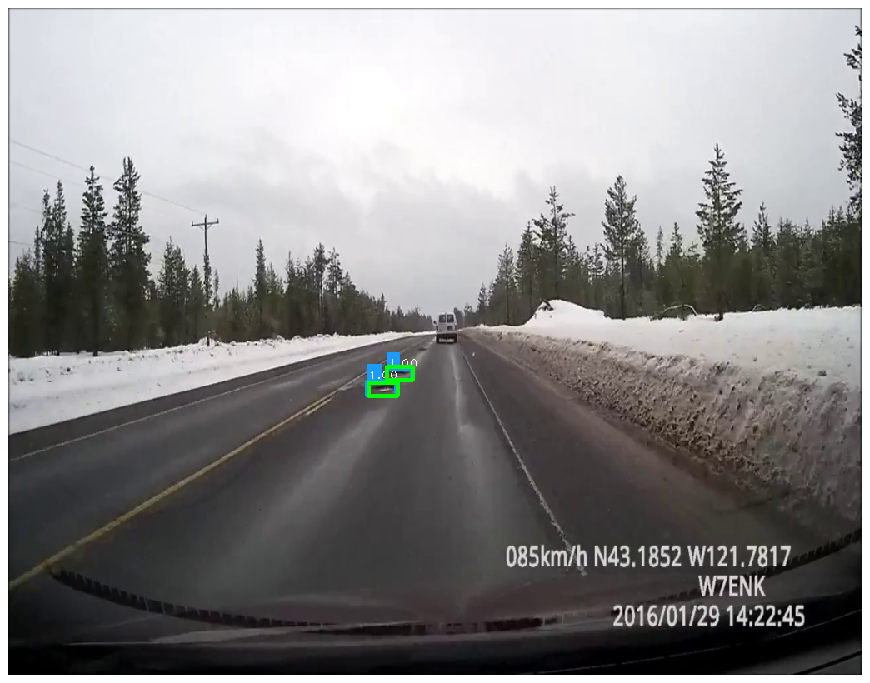}%
\hfil
\includegraphics[width=0.24\textwidth]{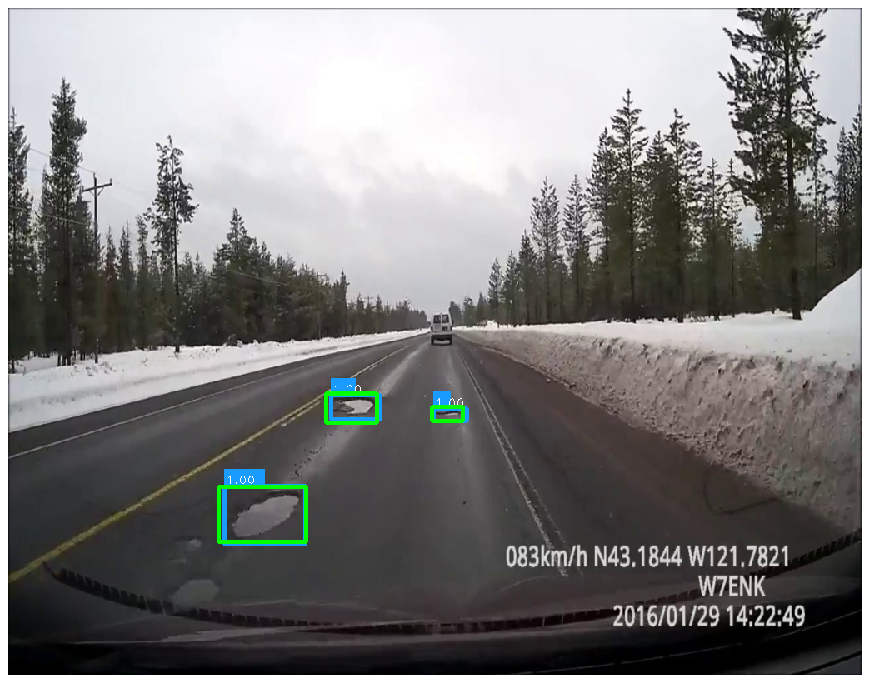}\\
\vspace{0.1cm}
\includegraphics[width=0.24\textwidth]{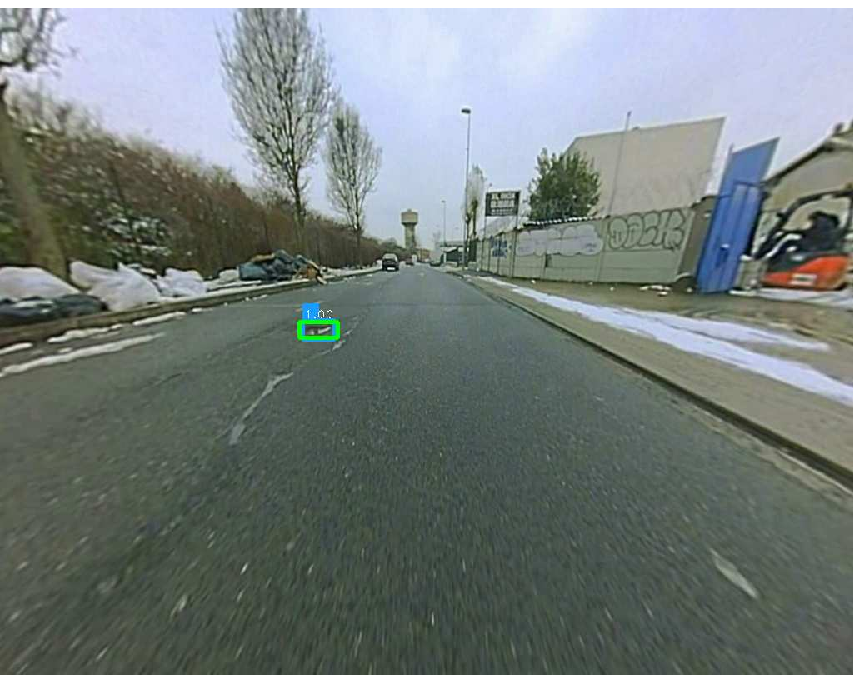}%
\hfil
\includegraphics[width=0.24\textwidth]{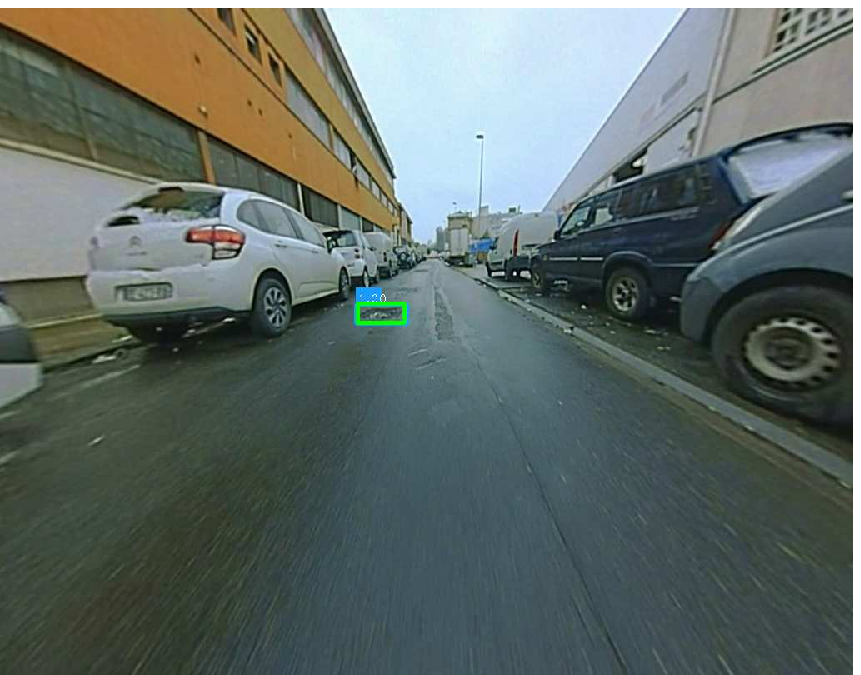}%
\hfil
\includegraphics[width=0.24\textwidth]{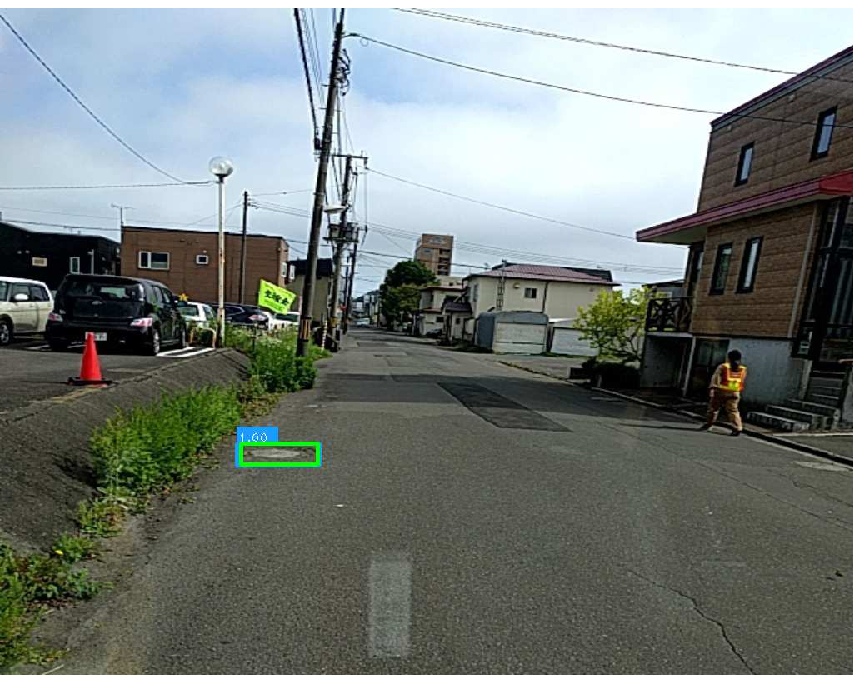}%
\hfil
\includegraphics[width=0.24\textwidth]{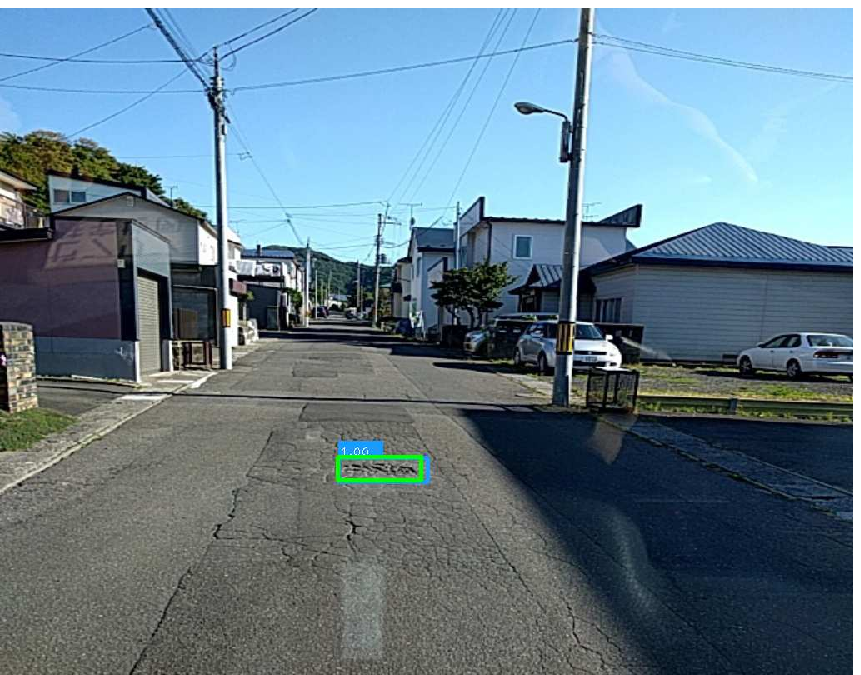}\\
\vspace{0.1cm}
\fcolorbox{green}{white}{\includegraphics[width=0.2\textwidth]{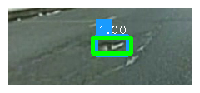}}%
\hfil
\fcolorbox{green}{white}{\includegraphics[width=0.2\textwidth]{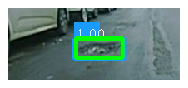}}%
\hfil
\fcolorbox{green}{white}{\includegraphics[width=0.2\textwidth]{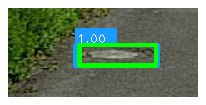}}%
\hfil
\fcolorbox{green}{white}{\includegraphics[width=0.2\textwidth]{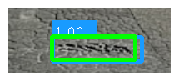}}\\
\caption{Correct pothole detections in the test set using the model \emph{Faster R-CNN Resnet101}. The ground-truth boxes are in green colour and the detections with overlaid score are in blue. The first row shows multiple detections, the second row are single potholes that are zoomed in and displayed in the third row.}
\label{fig:tpimages}
\end{figure*}

\begin{figure*}[!t]
\centering
\subfloat[Patch/pothole not annotated]{\includegraphics[width=0.24\textwidth]{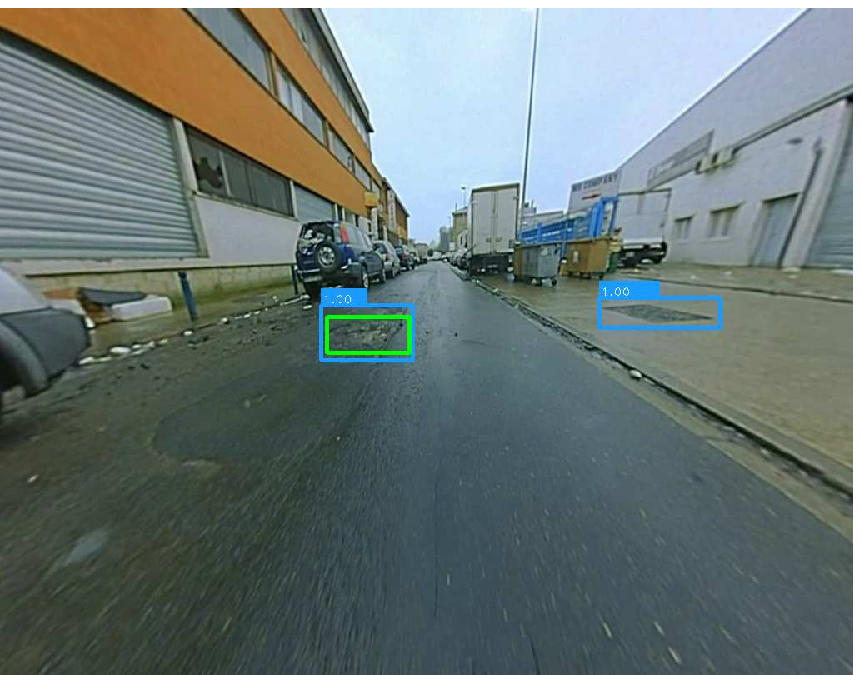}}%
\hfil
\subfloat[Oil smear similar to pothole]{\includegraphics[width=0.24\textwidth]{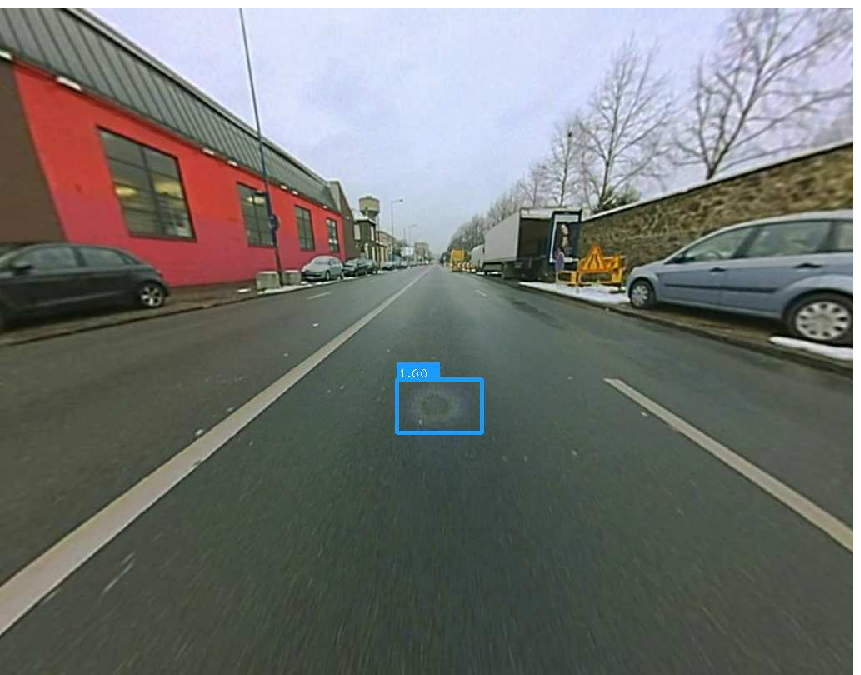}}%
\hfil
\subfloat[Water puddle at far distance]{\includegraphics[width=0.24\textwidth]{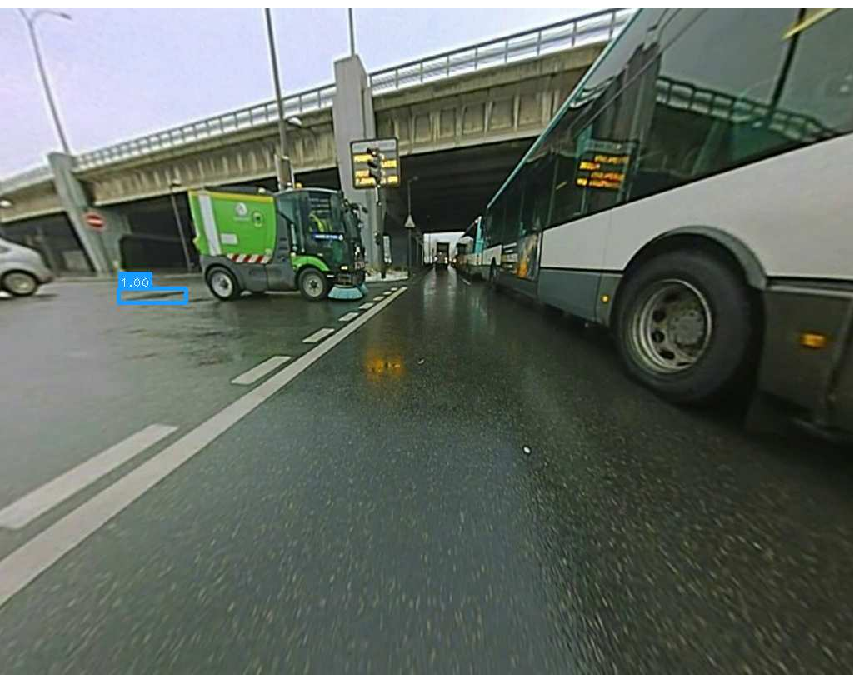}}%
\hfil
\subfloat[Stain on road surface]{\includegraphics[width=0.24\textwidth]{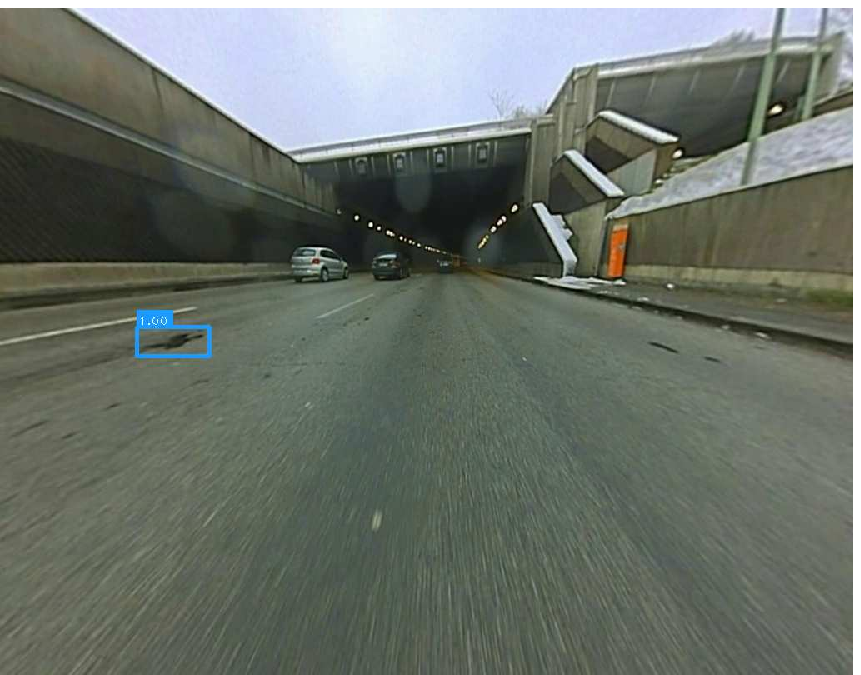}}\\
\subfloat[Correct detection but bigger than ground truth due to wet surface]{\includegraphics[width=0.24\textwidth]{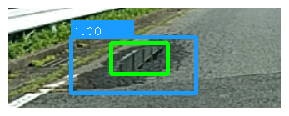}}%
\hfil
\subfloat[Small manhole within shallow road patch]{\includegraphics[width=0.2\textwidth]{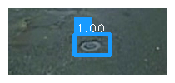}}%
\hspace{0.5cm}
\subfloat[Manhole with shallow border]{\includegraphics[width=0.24\textwidth]{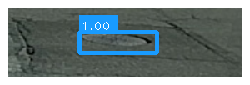}}%
\hfil
\subfloat[Manhole within shallow patch]{\includegraphics[width=0.24\textwidth]{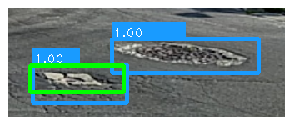}}\\
\caption{Examples of false positive detections in the test images using the model \emph{Faster R-CNN Resnet101}.}
\label{fig:fpimages}
\end{figure*}

\begin{figure*}[!t]
\centering
\subfloat[FN in green, but correctly detected in adjacent frames of the sequence]{\includegraphics[width=0.24\textwidth]{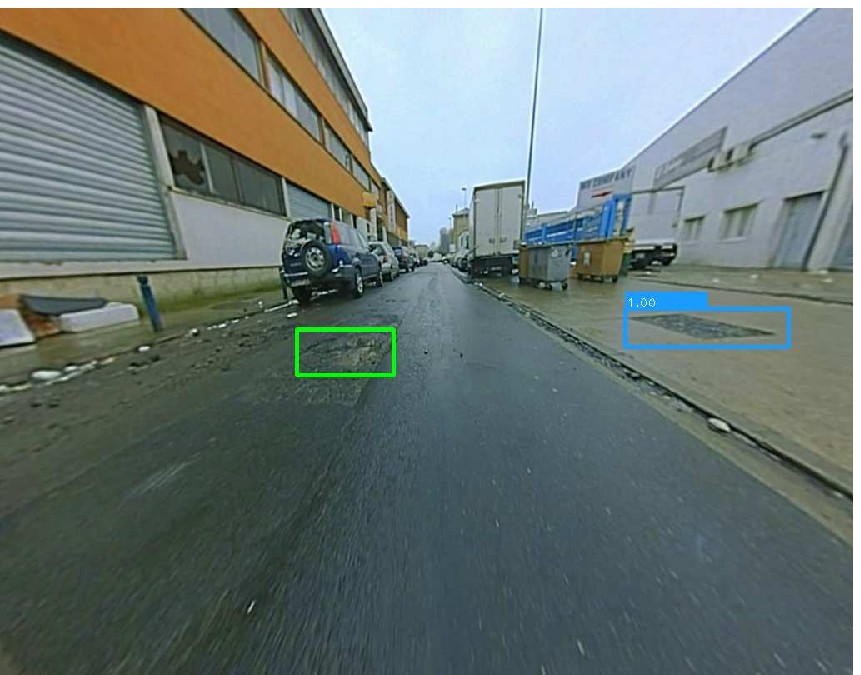}}%
\hfil
\subfloat[FN in green and TN manhole both not detected as potholes]{\includegraphics[width=0.24\textwidth]{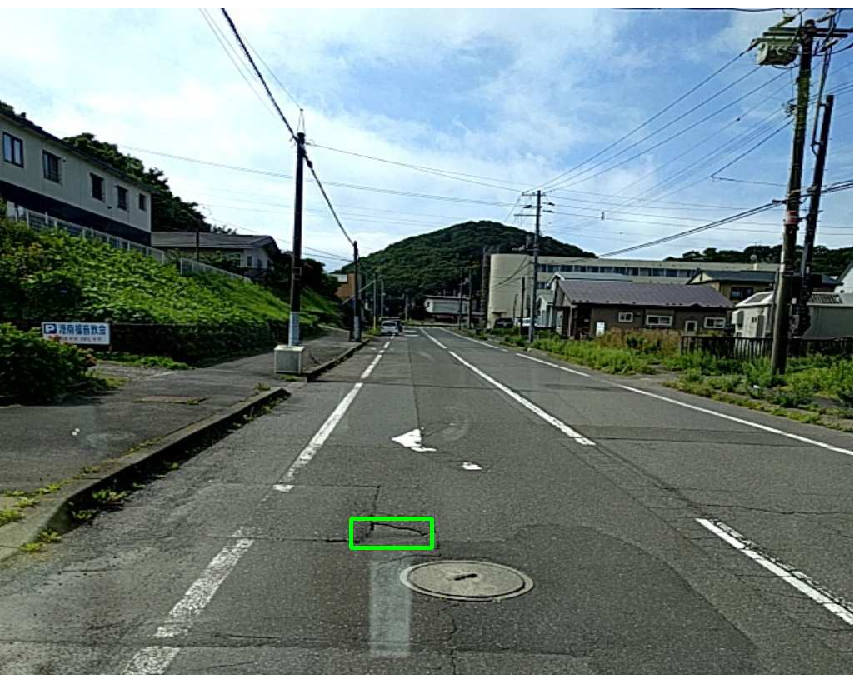}}%
\hfil
\subfloat[Surface stains inside tunnel and not detected as expected]{\includegraphics[width=0.24\textwidth]{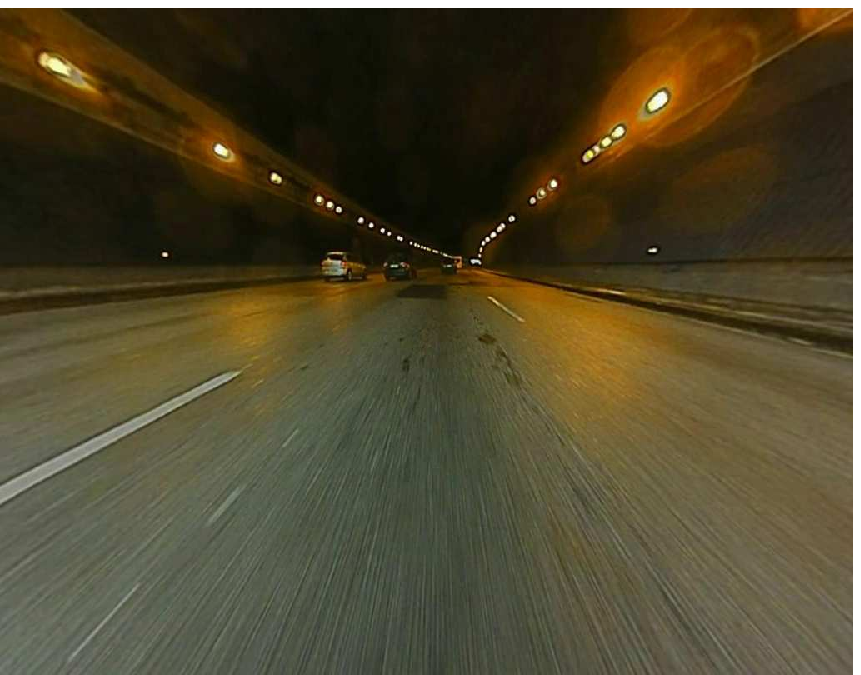}}%
\hfil
\subfloat[Correct behaviour not detecting manhole]{\includegraphics[width=0.24\textwidth]{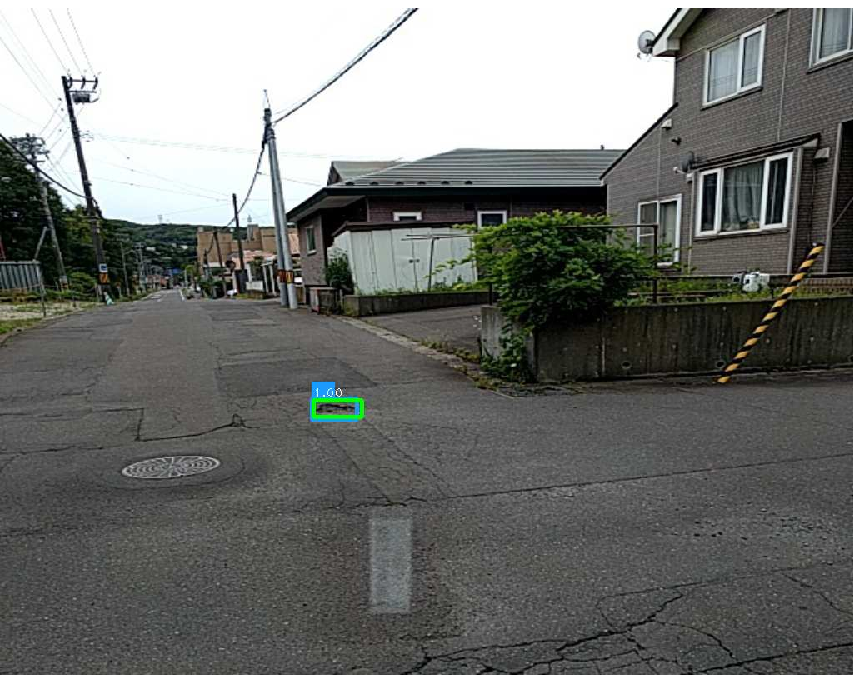}}\\
\caption{Examples of true and false negatives in the test images using the model \emph{Faster R-CNN Resnet101}.}
\label{fig:negimages}
\end{figure*}

Figures~\ref{fig:tpimages} to \ref{fig:negimages} depict some samples of the pothole detections with the best performing model, which is \textbf{Faster R-CNN Resnet101}. The potholes were correctly found in the scenes without any road surface filtering or manual pre-selection of image region of interest. The captions on the images describe the details of every case. We provide a supplementary video to demonstrate the achieved detection performance on a real world scene.

\section{Conclusion}\label{sec:conclusion}

This paper has presented how to automatically catch road potholes in worldwide real scenes. We built an image dataset with high intraclass variance from several sources with the aim of learning robust and general pothole appearance models that can aid the detection of this type of road hazards. We fine-tuned Deep Neural Networks for object detection with demonstrated good performance in the state of the art. We validated the suitability of some of these models for the particular goal of road pothole detection. We concluded that precise localization of potholes is challenging due to ground-truth errors during annotation because of the nature of road potholes. Thus, we relaxed the Jaccard index to $IoU=0.4$ for the evaluation. Consequently, \emph{Faster R-CNN Resnet101} achieved averaged precision values of 82\%, while \emph{Faster R-CNN Inception v2} yielded 75\% at a lower processing cost. The latter one, when tested on Nvidia DrivePX2 Autochauffeur platform, it could run at 5-6fps. Furthermore, the pothole detector was deployed in a real vehicle as part of AUTOPILOT project.

One of the limitations of the detector is the runtime in case of real-time requirements on the target system. For AUTOPILOT, the current frame rate is valid because potholes will be notified to IoT platforms upon detection and they will later serve road hazard warnings, hence, immediate reactive vehicle manoeuvres are not targeted. In addition, we have observed that most of the false positives are due to manholes. Sometimes, they are in a shallow road patch which could be considered as correct detection. Other false detections outside road boundaries could be easily filtered by using lane/road detectors. Moreover, in highly damaged road surfaces the detector found unlabelled potholes and some road cracks, which could be considered as correct due to the bad condition of the pavement. Also, a limitation of the algorithm is high speed driving ($>$60Km/h). Images show blur that filters edges and fine-grained details making very difficult the detection task. This limitation comes from the acquisition system, camera settings, illumination conditions and other related factors.

For the future we plan DNN optimizations to accelerate inference. One possibility is to split specific parts of the neural network in to CPU or GPU processing. Also, the evaluation of other lightweight networks that can be trained from scratch on the pothole dataset or the reduction of layers in state-of-the-art networks. For instance, YOLOv3~\cite{ref:Redmon2018ArxivYOLOv3} has recently reported state-of-the-art accuracy at faster inference time. For fine-tunning during learning, enlarging the dataset can help improve the detection ratios. Additionally, other road hazards of interest will be studied in the future as part of the AUTOPILOT project.

\section*{Acknowledgement}
This work has received funding from the European Research Council (ERC) under the European Union's H2020 research and innovation programme (grant agreement no. 731993, project AUTOPILOT).
The authors would like to thank Valeo Comfort and Driving Assistance (VCDA) SAS for the provision of a equipped vehicle and video recordings as part of AUTOPILOT. Also, the authors would like to thank Dr. M. Nieto and Dr. S. Sanchez that helped with the proofreading. 

\ifCLASSOPTIONcaptionsoff
  \newpage
\fi



\bibliographystyle{IEEEtran}
\bibliography{IEEEabrv,references}

%

\begin{IEEEbiography}[{\includegraphics[width=1in,height=1.25in,keepaspectratio]{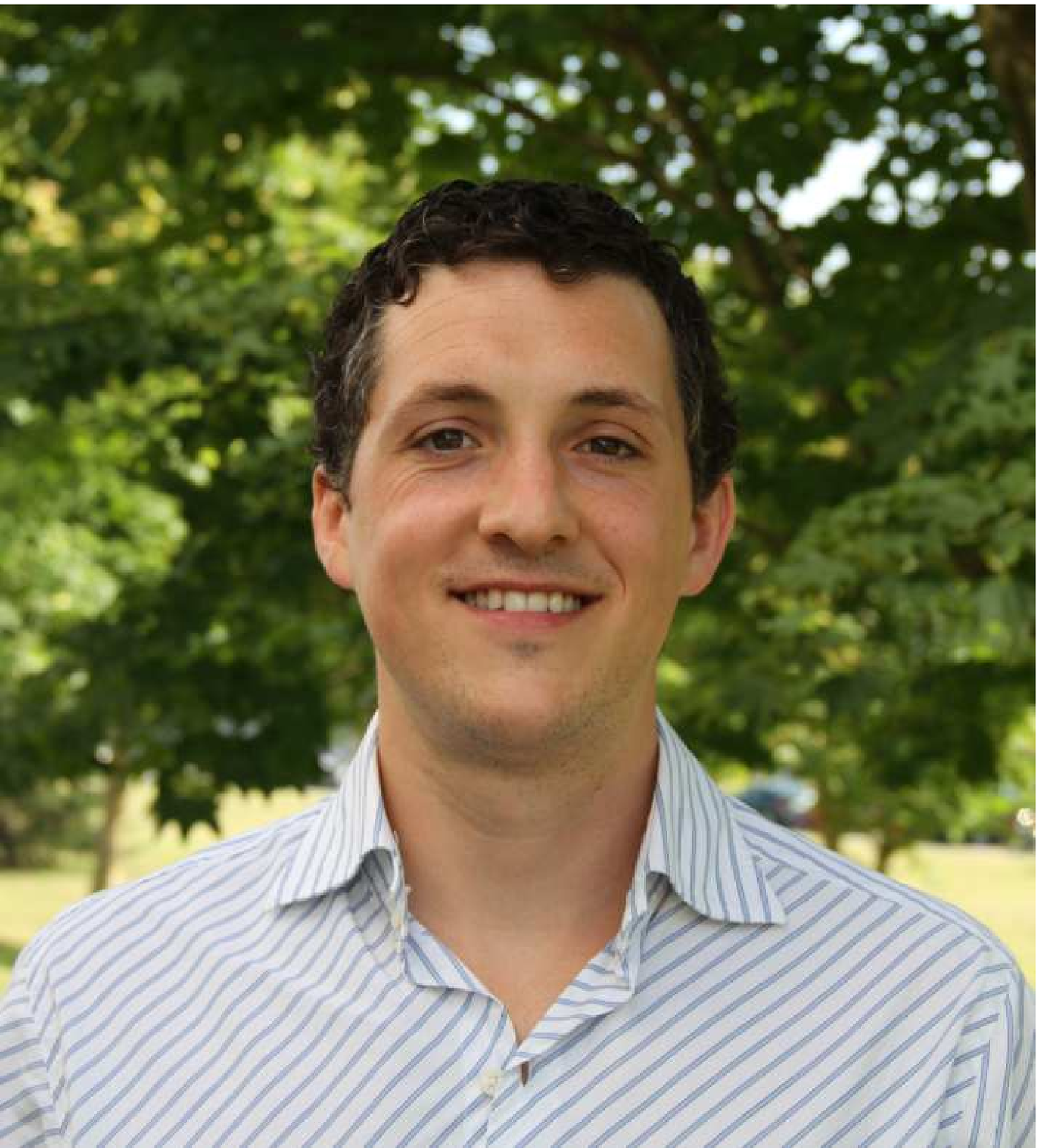}}]{J. Javier Yebes} received the Ph.D. degree (Cum Laude) in the field of Computer Vision and Robotics from the University of Alcal\'a (UAH), Spain, in 2014. He is a Researcher in computer vision and machine learning at Vicomtech, in the Department of ITS and Engineering, where he has a technical leading role for European H2020 projects. Also, he was the technical leader of an R\&D project for railways industry at Gobotix Ltd. (UK) in 2015-2017. He was pre- and post-Doctoral assistant at RobeSafe research group, UAH, from 2009 to 2015. His research interests include computer vision, deep learning, autonomous vehicles, ADAS, IoT and cloud.
\end{IEEEbiography}

\begin{IEEEbiography}[{\includegraphics[width=1in,height=1.25in,keepaspectratio]{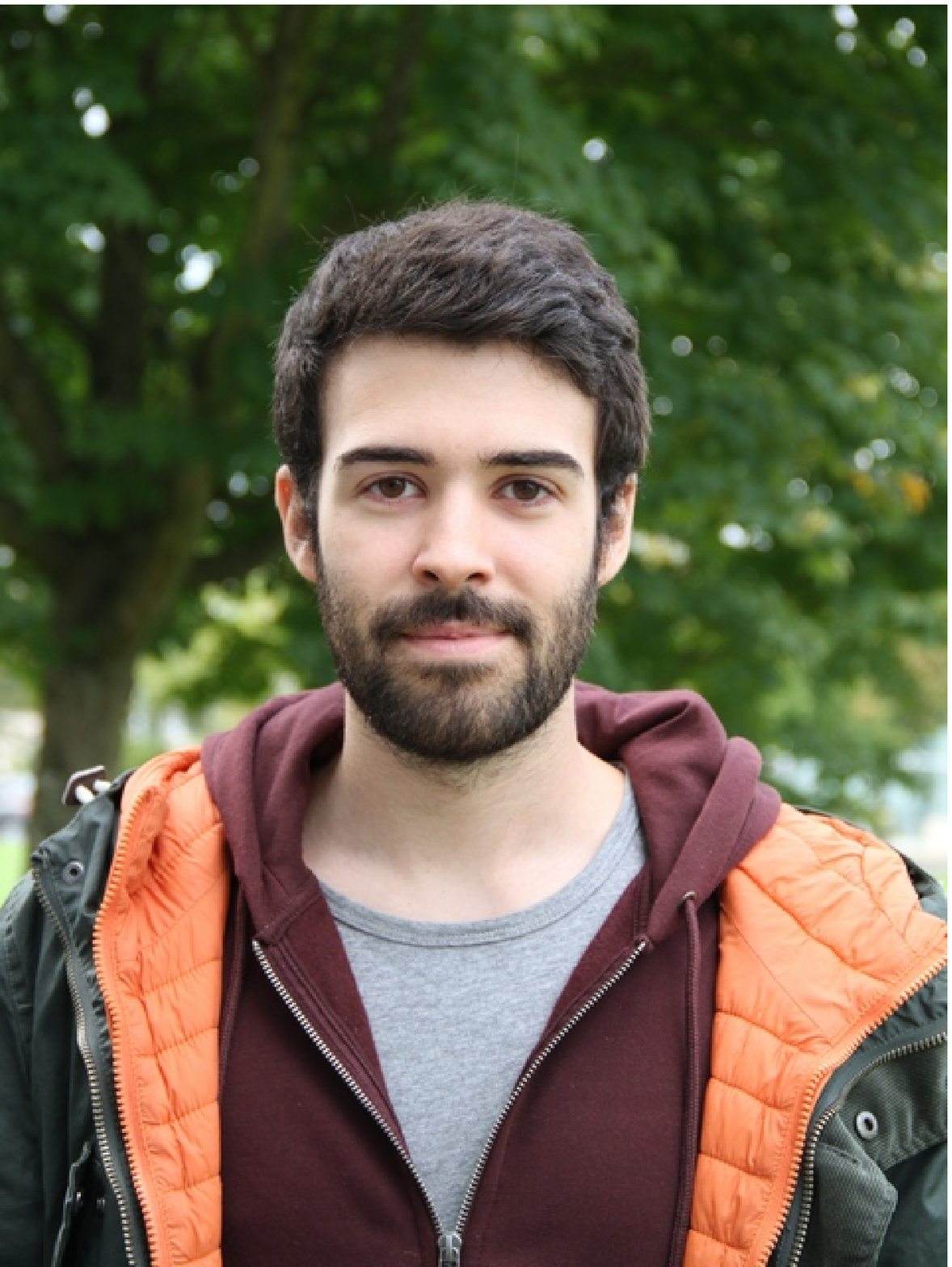}}]{David Montero} received the M.Sc. in Industrial Engineering in 2017 and the B.Sc. in 2014 from the University of Seville, Spain. He is a Researcher at Vicomtech, in the Department of ITS and Engineering where he is working in several private and public funded projects. During his studies, he worked on SLAM, 3D reconstruction and obstacle avoidance for UAVs. His research interests include computer vision and deep learning applied to autonomous vehicles and drones.
\end{IEEEbiography}

\begin{IEEEbiography}[{\includegraphics[width=1in,height=1.25in,keepaspectratio]{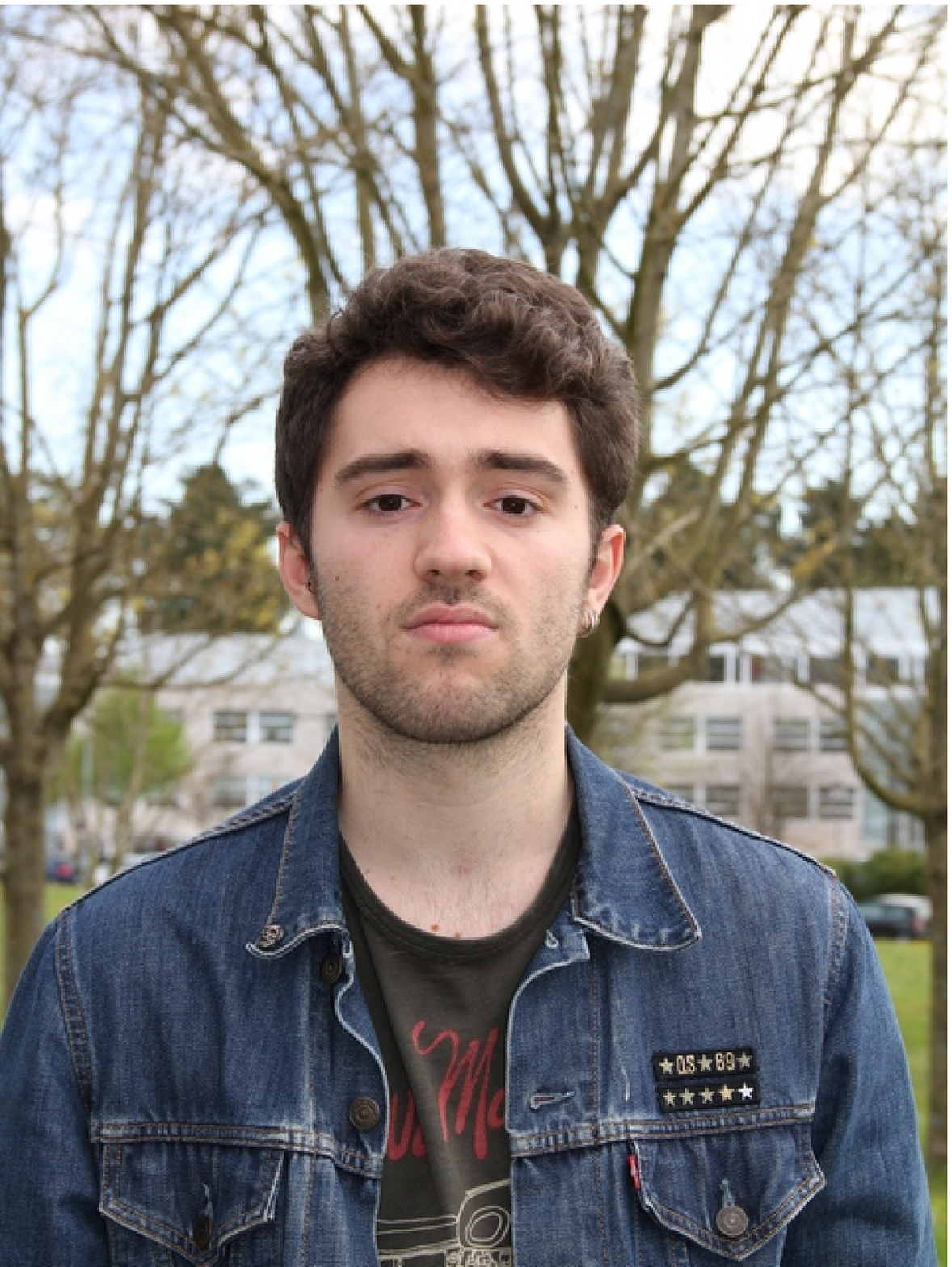}}]{Ignacio Arriola} received the M.Sc. in Computational Engineering and Intelligent Systems in 2018 and the B.Sc. in Electrical Engineering in 2017 from the University of Pais Vasco (UPV/EHU), Spain. During his degree thesis, he investigated the identification of road type based on driving patterns. He is currently an Assistant Researcher on computer vision and deep learning at Vicomtech. His research interests are in those fields applied to autonomous vehicles and ITS.
\end{IEEEbiography}




\end{document}